\documentclass[conference]{IEEEtran}
\usepackage{times}

\usepackage[numbers]{natbib}
\usepackage{multicol}
\usepackage[bookmarks=true]{hyperref}

\usepackage{amsmath}
\usepackage[utf8]{inputenc}
\usepackage{url}
\usepackage{booktabs}
\usepackage{amssymb}
\usepackage{bbding}
\usepackage{pifont}
\usepackage{wasysym}
\usepackage{utfsym}
\usepackage{fontawesome}

\usepackage{graphicx}
\usepackage{subcaption}

\begin{document}

\title{Advancing Humanoid Locomotion: Mastering Challenging Terrains with Denoising World Model Learning}

% % You will get a Paper-ID when submitting a pdf file to the conference system
% \author{Xinyang Gu$^{2*}$, Yen-Jen Wang$^{1,3*}$, Xiang Zhu$^{1,3*}$, Chengming Shi$^{1,3*}$, Yanjiang Guo$^{1,3*}$, Yichen Liu$^{1,3*}$, Jianyu Chen$^{1,2,3*}$
% }

\author{

\authorblockN{Xinyang Gu$^*$}
\authorblockA{RobotEra TECHNOLOGY CO., LTD.}
\authorblockA{$^*$Equal contribution. Project Co-lead.}
% \authorblockA{Email:xinyang.gu@robotera.com}

\and

\authorblockN{Yen-Jen Wang$^*$}
\authorblockA{Tsinghua University}
\authorblockA{Shanghai Qi Zhi Institute}
\authorblockA{Email: wangyenjen@berkeley.edu}
\authorblockA{$^*$Equal contribution. Project Co-lead.}

\and

\authorblockN{Xiang Zhu$^*$}
\authorblockA{Tsinghua University}
\authorblockA{Shanghai Qi Zhi Institute}
\authorblockA{$^*$Equal contribution. Project Co-lead.}
% \authorblockA{Email: zhu-x21@mails.tsinghua.edu.cn}

\and

\authorblockN{Chengming Shi$^*$}
\authorblockA{Tsinghua University}
\authorblockA{$^*$Equal contribution. Project Co-lead.}
% \authorblockA{Email: scm23@mails.tsinghua.edu.cn}

\and

\authorblockN{Yanjiang Guo}
\authorblockA{Tsinghua University}
\authorblockA{Shanghai Qi Zhi Institute}
% \authorblockA{Email: guoyj22@mails.tsinghua.edu.cn}

\and

\authorblockN{Yichen Liu}
\authorblockA{Tsinghua University}
% \authorblockA{Email: liu-yc22@mails.tsinghua.edu.cn}

\and

\authorblockN{Jianyu Chen$^\text{†}$}
\authorblockA{Tsinghua University}
\authorblockA{Shanghai Qi Zhi Institute}
\authorblockA{RobotEra TECHNOLOGY CO., LTD.}
\authorblockA{Email: jianyuchen@tsinghua.edu.cn}
\authorblockA{† Corresponding Author.}

}

% avoiding spaces at the end of the author lines is not a problem with
% conference papers because we don't use \thanks or \IEEEmembership

%Homer Simpson\authorrefmark{2},
%Montgomery Scott\authorrefmark{3} and
%Eldon Tyrell\authorrefmark{4}}
%\authorblockA  {\authorrefmark{3}Starfleet Academy, San Francisco, California 96678-2391\\
%Telephone: (800) 555--1212, Fax: (888) 555--1212}
%\authorblockA{\authorrefmark{4}Tyrell Inc., 123 Replicant Street, Los Angeles, California 90210--4321}}

\twocolumn[{%
\renewcommand\twocolumn[1][]{#1}%
\maketitle
\begin{center}
    \centering
    \hspace*{-1mm} 
    \includegraphics[width=1.002\linewidth]{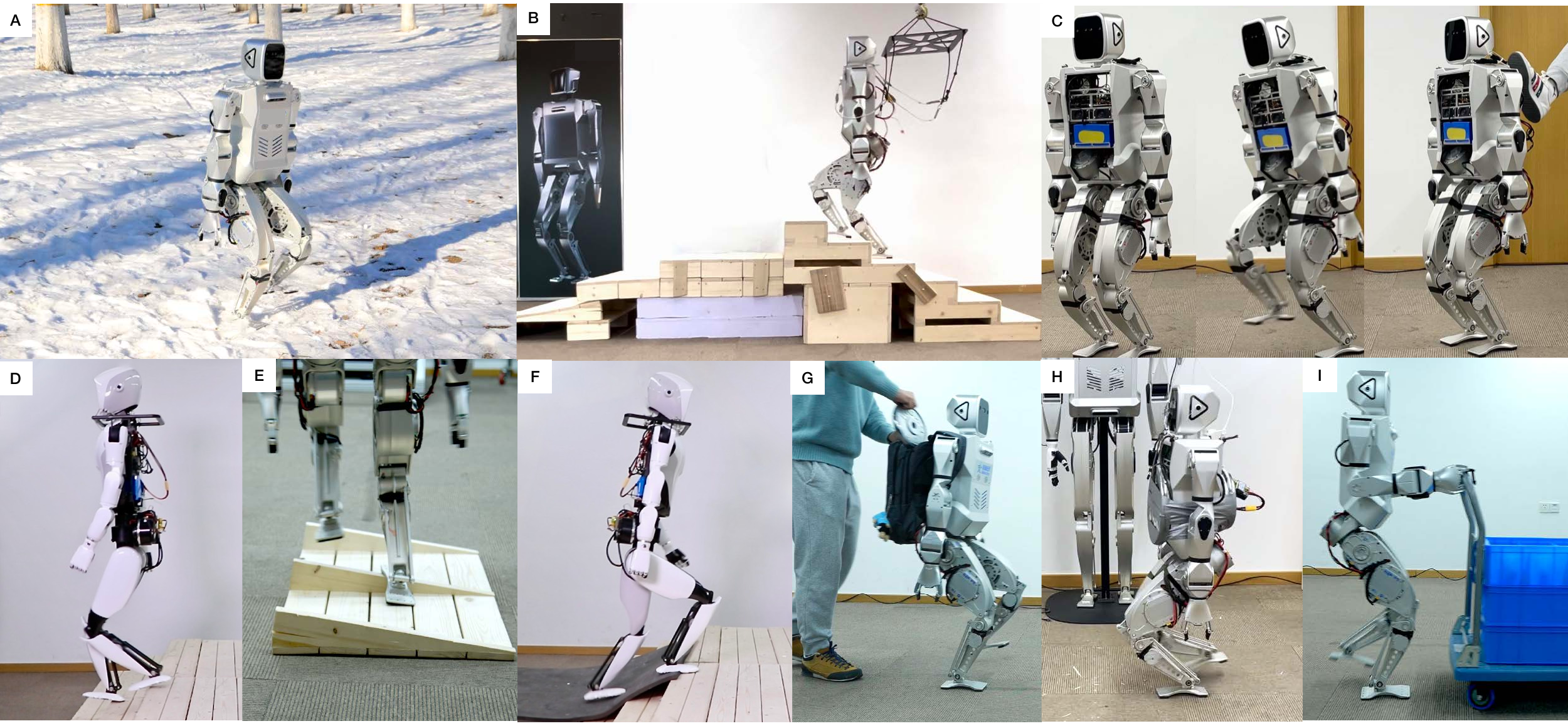}
    % \label{fig:main_exp}
    \vspace*{-5mm}
    \captionof{figure}{Extensive showcase of locomotion skills using the proposed framework. Displayed is a sequence illustrating a humanoid robot skillfully executing various locomotion tasks in real world challenging environments.}
    \label{fig:main_exp}
\end{center}%
% \vspace{-0.05in}
}]

\begin{abstract}
Humanoid robots, with their human-like skeletal structure, are especially suited for tasks in human-centric environments. However, this structure is accompanied by additional challenges in locomotion controller design, especially in complex real-world environments. As a result, existing humanoid robots are limited to relatively simple terrains, either with model-based control or model-free reinforcement learning. In this work, we introduce Denoising World Model Learning (DWL), an end-to-end reinforcement learning framework for humanoid locomotion control, which demonstrates the world's first humanoid robot to master real-world challenging terrains such as snowy and inclined land in the wild, up and down stairs, and extremely uneven terrains. All scenarios run the same learned neural network with zero-shot sim-to-real transfer, indicating the superior robustness and generalization capability of the proposed method.

\end{abstract}

% \begin{figure*}[tp]
%     \centering
%     \includegraphics[width=1\linewidth]{images/RSS_plot_main.pdf}
%     \caption{Comprehensive Locomotion Demonstrations Using the DWL Framework. This series portrays a humanoid robot adeptly undertaking varied locomotion tasks with the DWL framework across challenging settings, such as navigating through snow (A), maneuvering up and down stairs (B, D), traversing irregular terrains (H), and enduring additional loads and dynamic forces (E, F, G), demonstrating the framework's advanced capability for adaptive and robust locomotion.}
%     \label{fig:main_exp}
%     \vspace{-0.15in}
% \end{figure*}

\IEEEpeerreviewmaketitle

\section{Introduction}

% 性能 real-world challenge terrain first end-to-end
% pure RL;achievement
% challenge terrain
% contribution: clear and strong

% intro 思路： 分成四个部分ABCD
% A. humanoid important

% B. classical control ZMP; MPC
% leak the ability of challenge terrain locomotion; optimization is hard

% C. RL is good for this purpose
% Cassie; robot dog show potential
% but humanoid is more harder
% bike not on challenge terrain

% D. Our work
% Achievement
% propose DWL first humanoid robot that masters challenge terrain
% snow, stair pure end-to-end RL, in Fig2
% zero shot sim2real transfer, same policy without fine-tune
% why we can do it: DWL (world model) only TS learn action; single phase training process; SRL
% ankle; passive ankle or point foot; cassie one dim; first Parallel Double-Linkage Ankle Mechanism and why that is important; robust

% 近期，具備高動態及高泛化能力的機器人成為了一個主流的研究方向，包括

% A. humanoid important
% Given that modern environments are primarily designed for humans, a robot capable of maneuvering through complex terrains is crucial for real-world applications. Traditional mobile robotic platforms often face substantial limitations in such environments. In contrast, humanoid robots, emulating the human body's structure, have the inherent potential to traverse varied landscapes and perform tasks like humans.

Modern environments are primarily designed for humans. Therefore, humanoid robots, with their human-like skeletal structure, are especially suited for tasks in human-centric environments and offer unique advantages over other types of robots. Their mobility is crucial for completing diverse tasks in the real world, underlining the necessity of their capacity to walk on complex terrains.

% B. classical control ZMP; MPC

Previously, model-based control techniques such as Zero Moment Point (ZMP) and Model Predictive Control (MPC) combined with Whole-Body Control (WBC) have significantly advanced humanoid robots' locomotion abilities, enabling skills like walking, jumping, and even backflipping\cite{ahn2023development,wensing2014development,chignoli2021humanoid}. However, the success of these methods depends on accurately modeling the environment's dynamics, which can make it difficult to handle complex interactions with the environment, such as navigating challenging terrains. 

% ZMP maintains stability by ensuring the robot's center of pressure remains within the support area of its feet, preventing falls. Meanwhile, MPC with WBC predict the robot's immediate future movements and make adjustments to ensure the movements are smooth. 

% And a en

% C. RL is good for this purpose
Reinforcement learning (RL), on the other hand, relies less on exact environmental modeling. Recent progress in model-free RL has shown great potential, particularly in developing adaptive legged locomotion controllers \cite{rudin2022learning}. This allows robots to learn and adapt to a wide range of situations, often surpassing the capabilities of traditional model-based control methods\cite{li2023robust}.

However, ensuring robustness in humanoid robots, as opposed to quadrupedal\cite{rudin2022learning} and bipedal\cite{reher2019dynamic} counterparts, involves addressing several additional challenges. These include but are not limited to a higher center of gravity, instability during leg swinging, greater leg inertia, extra weight from the torso and arms, and their larger size overall. Therefore, to date, real-world applications of reinforcement learning (RL) for controlling humanoid robots, as demonstrated in the recent study\cite{radosavovic2023learning}, have been limited to relatively simple terrains.

% However, unlike bipedal \cite{reher2019dynamic} or quadruped robots \cite{rudin2022learning}, ensuring robustness as a higher center of gravity, instability during the swing phase, increased inertia in their legs, additional weight from the torso and arms, and their larger overall size. For now, only \cite{radosavovic2023learning} success used an RL for humanoid control in the real world, but rather on very simple terrain, mostly terrain. 

% In RL, robots are trained in a simulated environment to optimize a reward function, with the resulting policy then applied to the real robot. Despite its potential, RL faces several challenges, including bridging the significant gap between simulation and real-world physics, crafting appropriate reward functions for human-like walking, and the limited generality of RL algorithms, which may fail on atypical terrains.

% D. Our work
% Achievement
% propose DWL first humanoid robot that masters challenge terrain
% snow, stair pure end-to-end RL, in Fig2
% zero shot sim2real transfer, same policy without fine-tune
% why we can do it: DWL (world model) only TS learn action; single phase training process; SRL
% ankle; passive ankle or point foot; cassie one dim; first Parallel Double-Linkage Ankle Mechanism and why that is important; robust

\begin{figure}[tp]
    \centering
    \includegraphics[width=1.0\linewidth]{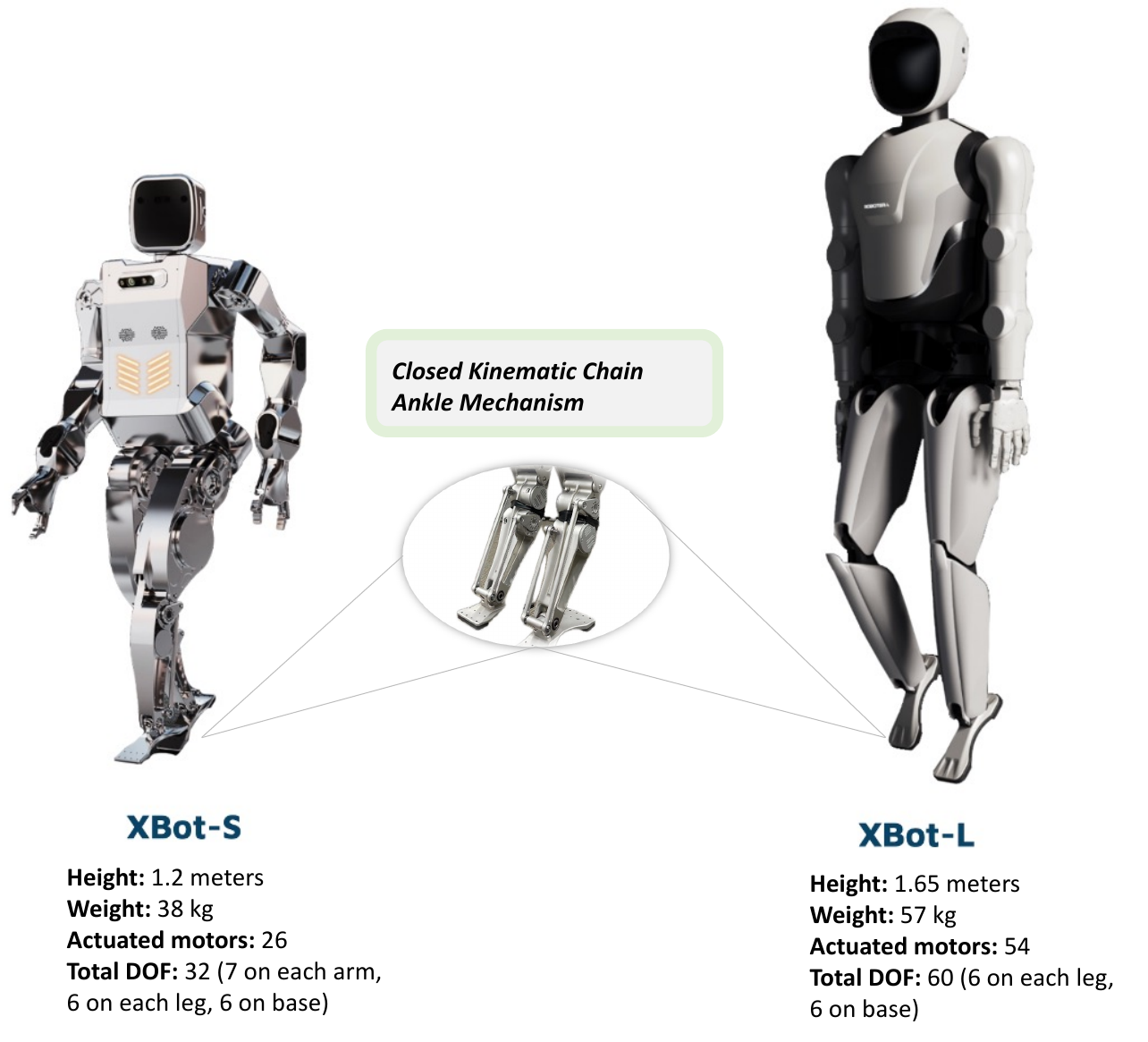}
    \caption{Illustration of the humanoid robot's hardware structure and the Closed Kinematic Chain Ankle Mechanism. This mechanism is notable for offering two degrees of freedom in each ankle while reducing leg inertia. Our works are tested on two distinct sizes of humanoid robots, XBot-S and XBot-L, provided by Robot Era.}
    \label{fig:ankle}
    \vspace{-0.15in}
\end{figure}

In this work, we introduce \textbf{Denoising World Model Learning (DWL)} for controlling humanoid robots across varied and complex terrains. To the best of our knowledge, DWL enables the world's first humanoid robot to master real-world challenging terrains with end-to-end RL and zero-shot sim-to-real transfer. As shown in Fig.\ref{fig:main_exp}, our humanoid robot is able to navigate stably through snowy inclined land in the wild, stairs, irregular surfaces, etc., and can resist large external disturbances. All scenarios run the same learned neural network policy, indicating its robustness and generalization. The key ingredient of DWL lies in establishing an effective representation learning framework to denoise the factors enlarging the sim-to-real gap. Furthermore, we are the first to enable active 2-DoF ankle control with a Closed Kinematic Chain Ankle Mechanism (shown in Fig.\ref{fig:ankle}) for humanoid robot locomotion learning. Unlike previous studies \cite{siekmann2021blind} with only one DoF ankle control or passive ankle control \cite{radosavovic2023learning}, our approach enables the robot to become extremely robust. The contributions of our work are summarized as follows:
\begin{enumerate}
    \item Demonstrate the world's first humanoid robot mastering real-world challenging terrains with end-to-end RL through zero-shot sim-to-real transfer.
    \item Propose DWL, a novel RL framework to bridge the sim-to-real gap and achieve robust generalizable performance.
    % \item Demonstrate the first humanoid robot using active 2-Dof ankle control with a Closed Kinematic Chain Ankle Mechanism with RL, which substantially enhances the stability and flexibility of the robots.

    % \item Present a humanoid robot with 2-Dof ankle control using a closed kinematic chain and RL, enhancing stability and flexibility.

    \item Unveil a cutting-edge humanoid robot with active 2-Dof ankle control, powered by RL and a closed kinematic chain for enhanced stability and flexibility.
    
\end{enumerate}

\section{Related Works}
\paragraph{Learning Robot Locomotion}
Reinforcement learning has become more promising to enable robots to perform stable locomotion\cite{tan2018sim, hwangbo2019learning, lee2020learning}. Compared to previous RL efforts with quadrupedal robots \cite{rudin2022learning} and bipedal robots like Cassie\cite{li2021reinforcement, kumar2022adapting}, our focus on humanoid robots presents a significantly more challenging setup. Our proposed method excels in automating state representation learning \cite{lesort2018state}, mastering end-to-end learning for both prediction and adaptation and facilitating a seamless zero-shot transfer to real-world scenarios by effectively bridging the sim-to-real gap.

Furthermore, conventional approaches, often encompassing multi-stage training processes \cite{agarwal2023legged}, detailed reward designs \cite{yang2020learning}, or behavior cloning \cite{merel2018neural}, typically falter amidst the dynamic and unpredictable real-world scenarios. On the other hand, DWL instead integrates a world model within an encoder-decoder framework, employing a masking loss to predict the state from observations. 

% This design functions as an efficient information bottleneck \cite{tishby2000information, tishby2015deep}, drawing inspiration from masked auto-encoders \cite{he2022masked, feichtenhofer2022masked, germain2015made, zhang2022survey} and the world model learning in RL \cite{hafner2019learning, hafner2019dream, hafner2020mastering}. The outcome is a methodology that not only ensures meaningful but also sparse learned representations, significantly enhancing the model's real-world transferability and effectiveness.

\paragraph{Humanoid Robot Locomotion Control}
The evolution of humanoid locomotion began with early concepts and basic models, exemplified by WABOT-1 in the 1970s\cite{kato1974information}. Progress in sensors and control algorithms enhanced humanoid robots' stability and adaptability. model-based control techniques\cite{kuindersma2016optimization} like ZMP\cite{sugihara2002real}, MPC\cite{sleiman2021unified, li2023multi}, and WBC\cite{sentis2006whole} have significantly improved locomotive capabilities. Learning-based approaches, which are less reliant on precise dynamic models, offer better adaptability and robustness. Despite this, real-world applications of reinforcement learning for humanoid control, such as \cite{radosavovic2023learning,gu2024humanoid}, have been successful but limited to simpler terrains.

\section{Problem Setting}
% \subsection{RL Background}
% We formulate the locomotion control problem as a discrete-time Partially observable Markov decision process (MDP), which is represented by a tuple $\mathcal{M}=\left\langle \mathcal{S}, \mathcal{A}, T, \mathcal{O}, R, p_0, \gamma\right\rangle$, where $\mathcal{S,A}$ denote the state and action space, $T(\bf{s'|s,a})$ and $R(\bf{s,a})$ denote the dynamics and reward function, $p_0$ denotes the initial state distribution and $\gamma\in(0,1)$ denotes the discount factor.

% At each timestep $t$, the agent observes the environment state $\bf{s}_t$ and the policy outputs a distribution over the action space $\pi(\bf{a}|\bf{s}_t)$. The agent then executes the action $\bf{a}_t$ sampled from the policy, interacts with the environment and receives a new state $\bf{s}_{t+1}$ and a reward $r_t$. The objective of RL is to maximize the return, i.e. the expected accumulative discounted reward $J=\mathbb{E}\left[\sum_t\gamma^tr_t\right]$ over an episode.

\subsection{Reinforcement Learning Background}

Our approach utilizes a reinforcement learning problem setting, encapsulated in the tuple \(\mathcal{M} = \langle \mathcal{S}, \mathcal{A}, T, \mathcal{O}, R, \gamma \rangle\). Here, \(\mathcal{S}\) and \(\mathcal{A}\) represent the state and action spaces, with the transition dynamics \(T(\mathbf{s}'|\mathbf{s},\mathbf{a})\), the reward function \(R(\mathbf{s},\mathbf{a})\), and the discount factor \(\gamma \in [0, 1]\). \(\mathcal{O}\) represents the observation space. 

Our framework distinctly adapts to both simulated and real-world environments. In the simulation, the agent is afforded complete visibility of state $\mathbf{s}\in\mathcal{S}$. On the other hand, the real world is plagued by partial observability. The agent only has access to partially observations \(\mathbf{o}\in\mathcal{O}\), which provide incomplete information about the state due to sensory limitations and environmental noise. The policy \(\pi(\mathbf{a}|\mathbf{o}_{\leq t})\) maps the historical observations to a distribution over actions. As a result, the agent operates within a discrete-time Partially Observable Markov Decision Process (POMDP), necessitating decision-making based on sporadic and partial data. The primary goal is to optimize this policy \(\pi\) to maximize the expected total return \(J = \mathbb{E}[R_t] = \mathbb{E}\left[\sum_{t}\gamma^t r_t\right] \).

% In this context, the agent selects an action \(\mathbf{a}_t\) based on the current policy, which then leads to the transition to a new state \(\mathbf{s}_{t+1}\) and a corresponding reward \(r_t\). The primary goal in reinforcement learning is to optimize this policy \(\pi\) to maximize the expected total return \(J = \mathbb{E}[R_t] = \mathbb{E}\left[\sum_{t}\gamma^t r_t\right] \).
% which represents the cumulative discounted reward over an episode. 
% \begin{figure}[tp]
%     \centering
%     \includegraphics[width=0.7\linewidth]{images/ankle.pdf}
%     \caption{Illustration of the humanoid robot's overall structure and the intricately designed Parallel Double-Linkage Ankle Mechanism. This mechanism is notable for offering two degrees of freedom in each ankle, enabling enhanced mobility and stability.}
%     \label{fig:ankle}
%     \vspace{-0.15in}
% \end{figure}

\subsection{Humanoid Robot Hardware}

We use two different sizes of humanoid robots for our experiments, as illustrated in Fig.~\ref{fig:ankle}. XBot-S weighs 38 kg and stands 1.2 meters tall. The robot is equipped with 26 actuated motors: 7 in each arm and 6 in each leg. XBot-L weighs 57 kg and stands 1.65 meters tall. The robot is equipped with 54 actuated motors. For the purposes of this study, we focus on leg control, keeping the arm motors stationary. Each leg is powered by 6 motors: the yaw and roll joint motors with a peak torque of $100 \mathrm{N} \cdot \mathrm{m}$, the pitch and knee joint motors with $250 \mathrm{N} \cdot \mathrm{m}$, and 2 ankle motors each providing $36 \mathrm{N} \cdot \mathrm{m}$ of torque ($50 \mathrm{N} \cdot \mathrm{m}$ in XBot-L). The ankle motors, situated near the knee, are operated via a \emph{Closed Kinematic Chain Ankle Mechanism} as Fig.~\ref{fig:ankle}. This design aims to reduce leg inertia while ensuring an adequate degree of freedom.

% The observables consist of 2 parts: 1) The leg joint positions and velocities from onboard joint encoders and 2) The floating base rotational positions and velocities from IMUs. 

% The robot's construction employs advanced materials and actuation systems, facilitating precise control over its movements. The inclusion of the Parallel Double-Linkage Ankle Mechanism is a deliberate choice to replicate the complex motion patterns of human walking, which is critical for navigating uneven surfaces and maintaining postural stability. Each joint is driven by custom design high-torque actuators, allowing for smooth transitions between different motion states. Sensor arrays embedded throughout the robot provide real-time feedback for proproception, contributing to its autonomous decision-making capabilities.

\section{Methods}
\subsection{Denoising World Model Learning}
\label{sec:DWL}

Utilizing RL, various skills can be learned in simulation, but the transition to real-world robots faces significant challenges due to the sim-to-real gap, which is mainly caused by inaccurate simulation of the robot hardware and the limited information provided by onboard sensors. To overcome this barrier, we introduce \textbf{Denoising World Model Learning (DWL)}, which enables online adaptation and state estimation through representation learning. DWL is characterized by two primary features:

\begin{itemize}
    % \item An encoder-decoder mechanism dedicated to world model learning, with latent space embedding it reconstruction robot's full state from partially observed raw sensory data.

    \item An encoder-decoder architecture for world model learning, effectively embedding partially observed historical raw sensory data into a latent space and reconstructing the robot's full state from it.
    
    \item A policy gradient method that facilitates iterative improvements of the controller, allowing optimizing complex objectives through environmental interaction.
\end{itemize}

\subsubsection{Encoder-Decoder Architecture of DWL}

% If we could have a perfect simulator with fully-observable accurate sensors on the robot, then there would be no sim-to-real gap. But unluckily, we can only get noisy raw sensor data in the real world. The sim-to-real gap can be regarded as adding the following types of \textbf{noises} to the sensor observation:

% A fully observable state, accurate sensors, and a perfect simulator would eliminate the sim-to-real gap in an ideal world. However, real-world scenarios only provide noisy, partially observed sensor data, and the simulator is far from perfect. Thus, the sim-to-real gap can be regarded as adding the following types of \textbf{noises} to the true state:

In an ideal world, a complete observed state, precise sensors, and a flawless simulator would eliminate the sim-to-real gap. Yet, reality often presents us with noisy, incomplete sensor data, and the simulator is far from perfect. Consequently, the sim-to-real gap can be conceptualized as introducing the following types of \textbf{noises} to the true state:

% \begin{itemize}
%     \item Environment noise: there are uncertain complex environments and unexpected external disturbances in real world, such as navigating on challenging terrains or we apply a random push force to the robot.
%     \item Dynamics noise: simulating the exactly true dynamics of the physical world is intractable, so there are a lot of simplifications in the simulator, for example, the friction of the ground, or the deformability of the objects.
%     \item Sensory noise: all physical sensors have some measurement noise, such as the draft of IMU, and the joint position measurements.
%     \item Masking noise: There might some information that could not be directly obtained in the real world just because we simply don't have the corresponding sensors on the robot, such as the linear velocity and the contact force. This partial observability can be regarded as adding masking noise\cite{devlin2018bert, han2018masking, he2022masked}.
% \end{itemize}
\begin{itemize}
    \item Environmental noise: Real-world environments are complex and unpredictable, presenting challenges such as navigating on challenging terrains or unexpected external forces applied to the robot.
    \item Dynamics noise: Accurately simulating the true dynamics of the physical world is unfeasible, leading to oversimplifications in simulations, like approximations of ground friction or object deformability.
    \item Sensory noise: Physical sensors inherently contain measurement noise, for example, IMU drift and inaccuracies in joint position readings.
    \item Masking noise: Some information may be unobtainable in reality due to the absence of specific sensors on the robot, such as linear velocity and contact force measurements. This partial observability can be regarded as adding masking noise\cite{devlin2018bert, han2018masking, he2022masked}.
\end{itemize}

% To address these noises, we design a framework to firstly simulate the noisy observations in the simulation, and then denoise the observations
% to recover the true state and dynamics with an encoder-decoder architecture, as shown in Fig. \ref{fig:method}. 
To mitigate these noises, we have developed a framework that firstly simulates noisy observations within the simulation and subsequently employs an encoder-decoder architecture to denoise these observations and accurately recover the true state and dynamics, as depicted in Fig. \ref{fig:method}.

% Furthermore, we mask out some information that is not observable from the robot to simulate the masking due to partial observability. We then simulate the environment, dynamics, and sensor noises with domain randomization (DR) techniques\cite{tobin2017domain}, where we randomly perturbed the true state and dynamics such as the angular velocity and PD factors. This process could be described by an observation model\cite{hafner2019learning} $o_t\sim P_{\text{Noise}}(o_t | s_t)$. Details are provided in sec \ref{sec:DR}.

Additionally, to mimic the constraints of partial observability, we mask out some information that is not observable on real robots. We emulate the environment, dynamics, and sensor noises utilizing domain randomization (DR) methods\cite{tobin2017domain}. In this approach, we introduce random perturbations to the actual state and dynamics, such as angular velocity and PD parameters. This procedure aligns with an observation model\cite{hafner2019learning} expressed as $o_t\sim P_{\text{Noise}}(o_t | s_t)$. We elaborate on this in Section \ref{sec:DR}.

An encoder-decoder architecture is designed to denoise the observations. The recurrent encoder extracts latent state $z_t$ from the robot's historical noisy sensor observations. This latent representation is the core of state estimation, providing a rich, condensed summary of the robot's situational awareness. Subsequently, the decoder endeavors to reconstruct the robot's true state from this latent state.
% The learning paradigm is governed by a probabilistic model acting like a denoising process, which is pivotal in reconstructing state information from raw sensor data.
% The probabilistic model is rooted in a generative process, aiming to learn the underlying distribution of states that captures the robot's interaction with the environment, enabling the robot to infer and predict states even under the presence of sensory limitation and uncertainty. 
The formal expression of this model is given by:
\begin{equation}
    P(\tilde{\mathbf{s}}_{t}) = \mathrm{E}_{ o_{\leq t}} \left[ \int_z P_{\text{Decoder}}(\tilde{\mathbf{s}}_{t} | z_t) \cdot P_{\text{Encoder}}(z_t | o_{\leq t}) \right]
\end{equation}
where \( P(\tilde{\mathbf{s}}_{t}) \) represents the estimation of the real state distribution \( P(\mathbf{s}_t) \) at time \( t \). The encoder captures the conditional distribution \( P_{\text{Encoder}} \) of these latent variables given the noisy historical observations \( o_{\leq t} \), and the decoder \( P_{\text{Decoder}} \) reconstructs the state from the latent representation $z_t$.

It is imperative to recognize that the dimension of \(o_{\leq t}\) is 
much larger than that of $z_t$, implying \(z_t\) an effective information bottleneck \cite{tishby2000information}. This allows DWL to prioritize the salient aspects of sensory input. Furthermore, to enhance the efficiency and robustness of state estimation, sparsity within the latent representation is sought \cite{chen2022sparsity}. This is achieved by introducing an L1 regularization term in the latent domain. Moreover, since there is no need to generate new data from the latent space, and it is in a pure denoising process, a deterministic loss could be adopted instead of a variational loss \cite{kingma2013auto}. The denoising loss is thus expressed as follows:
\begin{equation}
\label{eq:denoise}
\mathcal{L}_{\mathrm{denoise}}=\| \tilde{\mathbf{s}}_{t} - \mathbf{s}_t \|_2+\lambda_r \|\boldsymbol{z}_t\|_1
\end{equation}
where \( \lambda_r \) represents the regularization coefficient. And $s_t$ is the full state. By incorporating privileged information into the state, such as the ground's friction coefficients, the actuator's torque values, and terrain height scans, enables the agent to effectively conduct online adaptation and system identification.

\subsubsection{Policy Learning in DWL}

Within the DWL framework, an Asymmetric Actor-Critic architecture is employed, drawing upon the concept of privilege learning as elucidated in previous works\cite{chen2020learning} \cite{pinto2017asymmetric}. This architecture is instrumental in enhancing data utilization during the training phase, proving particularly beneficial in real-world scenarios where direct state information is inaccessible. The actor component of the model computes its loss via the Proximal Policy Optimization (PPO)\cite{schulman2017proximal}, which is articulated as:
\begin{equation}
\label{eq:policy}
% \scriptsize
% \footnotesize
\begin{aligned}
\mathcal{L}_{\pi} &= \min \left[
\frac{\pi(a_t \mid o_{\leq t})}{\pi_{b}(a_t \mid o_{\leq t})} A^{\pi_{b}}(o_{\leq t}, a_t), \right. \\
&\quad \left. \text{clip}\left(\frac{\pi(a_t \mid o_{\leq t})}{\pi_{b}(a_t \mid o_{\leq t})}, c_1, c_2\right) A^{\pi_{b}}(o_{\leq t}, a_t)
\right]
\end{aligned}
\end{equation}
where \(\pi\) denotes the target policy to be optimized, \(\pi_{\text{b}}\) is the behavior policy employed for data sampling, and $c_1, c_2$ represents the PPO clipping range. In the context of DWL's encoder-decoder structure, the actor policy is defined as \(\pi(a_t \mid P_{\text{Encoder}}(z_t \mid o_{\leq t}))\). On the other hand, the critic could use state information for calculations of the value function. Thus, the critic loss is given by the following formula:

\begin{equation}
\label{eq:value}
\mathcal{L}_v = \| R_t - V(s_t)\|_2,
\end{equation}

In this formulation, \(R_t\) denotes the cumulative return at time \(t\), and \(V(s_t)\) is the value function as determined by the critic at state \(s_t\). The integration of privileged information within the state representation arms the learning agent with the capacity to make informed decisions. This approach aligns seamlessly with the DWL framework. It obviates the need for design privilege information by employing a unified state definition for both state estimation and value function assessment.

\subsubsection{Formulating the DWL Loss Function}

The DWL framework consolidates its learning objectives through a composite loss function that integrates the aspects of state reconstruction and policy optimization. The total loss function is a weighted sum of the denoising loss Equation(\ref{eq:denoise}), the policy loss Equation(\ref{eq:policy}), and the value loss Equation(\ref{eq:value}), formally expressed as:

\begin{equation}
\label{eq:dwl}
\mathcal{L}_{\text{DWL}} = \mathcal{L}_{\text{denoise}} + \lambda_{\pi} \mathcal{L}_{\pi} + \lambda_{v} \mathcal{L}_{v},
\end{equation}

% where \(\lambda_{\pi}\) and \(\lambda_{v}\) are the weighting factors that modulate the influence of the policy and value losses, respectively. This approach allows the DWL to adeptly calibrate the learning process, ensuring that the model is not only capable of accurate state estimation but also of making informed decisions based on these estimations. By reconstructing privilege information, DWL demonstrates robustness and adaptability in complex, real-world scenarios, as further evidenced in Section \ref{sec:exp}.

% In summary, through the integration of masked noise and domain randomization noise, we achieve a robust latent embedding. Coupled with policy gradient loss, our approach facilitates end-to-end State Representation Learning and policy optimization. This forms a sophisticated mechanism that efficiently bridges the gap between simulation and real-world application.

where \(\lambda_{\pi}\) and \(\lambda_{v}\) are the weighting factors for the policy and value loss components, respectively. This approach enables DWL to fine-tune the learning process, ensuring precise state estimation and informed decision-making based on these estimates. Reconstructing the state from masking loss and domain randomization noise, DWL demonstrates robustness and adaptability in complex real-world scenarios, as elaborated in Section \ref{sec:exp}.

In conclusion, the integration of masking noise and domain randomization noise cultivates a robust latent space for end-to-end state representation learning. When allied with policy gradient loss, this strategy propels a comprehensive approach to state estimation and policy optimization. This refined system is adept at narrowing the sim-to-real gap, effectively translating simulation-trained models to real-world applications.

% This is also related to the inability of our robot's foot linkage mechanism to simulate in simulation which consequently leads to a greater challenge in approximating the magnitude of this contact force with reality if big foot-ground forces are generated in simulation.
% If the robot continues to learn under such incorrect dynamics, it will result in a larger sim-to-real gap.
% (heuristic and regularization; \textbf{feet trajectory}; root trajectory)
\begin{figure*}[htp]
    \centering
    \includegraphics[width=1.\linewidth]{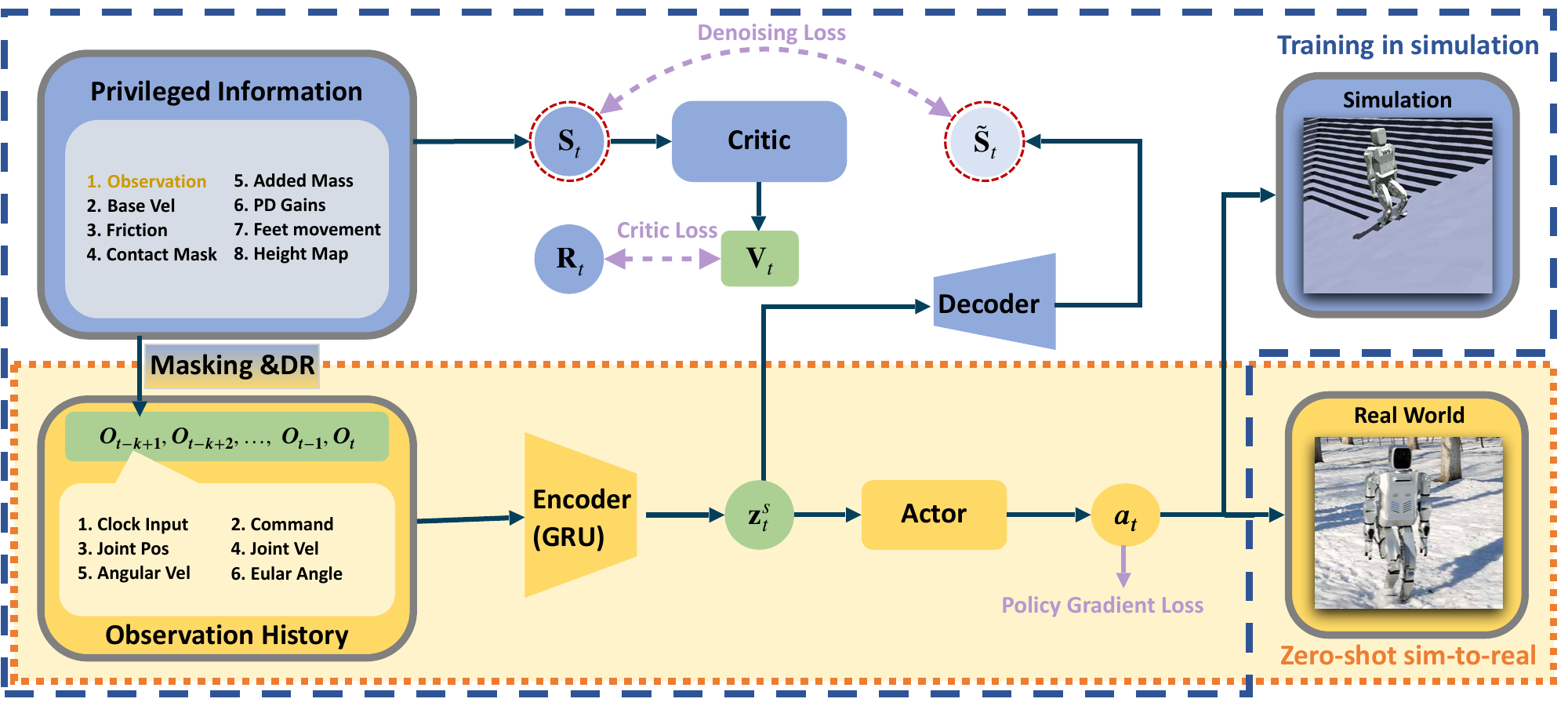}
    % \caption{Schematic of the Denoising World Model Learning framework. This illustration delineates the flow of information from sensory inputs to action outputs within both simulated and real-world environments. The raw observations are generated by adding masking and DR noises to the privileged information, which is encoded to a latent state, followed by a decoder to reconstruct the true state through a denoising process.} %It highlights the critical role of the encoder in processing the observation history, the subsequent policy learning, and the deployment of the predictive model for robust locomotion through the MLP decoder. The architecture underscores the integration of privileged information into the state representation, facilitating enhanced performance and generalization.}

    \caption{Illustration of the Denoising World Model Learning Framework. This diagram details the information flow from sensory input to action output in both simulated and real-world settings. Raw observations are generated by adding masking and DR noise to privileged observations. This is then encoded into a latent state and decoded to reconstruct the true state via a denoising process.}
    \label{fig:method}
    \vspace{-0.15in}
\end{figure*}

\subsection{Reward Formulation}
\label{sec:Reward}

Our reward function guides the robot to follow velocity commands, maintain a stable gait, and ensure gentle contact, thereby enabling robust locomotion across challenging terrains and above-ground obstacles.

\subsubsection{Composition of Rewards}

The reward function is structured into four key components: (1) velocity tracking, (2) periodic reward, (3) feet trajectory tracking, and (4) regularization terms. Our approach, drawing inspiration from previous works \cite{siekmann2021sim, radosavovic2023learning, siekmann2021blind}, employs a periodic reward to facilitate natural gait learning. Furthermore, we introduce a tracking loss defined as $\phi(e, w):=\exp \left(-w \cdot \|e\|^2\right)$, where \( e \) represents the tracking error and \( w \) the error tolerance strength. Detailed calculations can be found in the appendix, Section \ref{sec:reward_funciton}.

A novel aspect of our reward design addresses the sparse nature of contact force feedback. Rather than relying exclusively on contact force, our system focuses on foot velocity tracking. This is achieved by designing foot trajectories that incorporate predetermined velocities upon ground contact, thus ensuring a consistent and robust reward signal at each step. Such a strategy promotes gentle ground contacts, leading to reduced impact forces and enhancing the effectiveness of the sim-to-real transfer.

\subsubsection{Quintic Polynomial Foot Trajectory Interpolation}

In our approach, we focus on refining the locomotion of humanoid robots through the strategic design of foot trajectories. Quintic polynomial interpolation is utilized to determine these trajectories, a method that is particularly effective in meeting the precise kinematic requirements of a humanoid robot's gait cycle. This technique not only facilitates smoother motion but also ensures accurate foot placement, a crucial factor for maintaining stability and efficiency in humanoid walking patterns.

Quintic polynomial interpolation offers an advantageous approach in robotic motion planning due to its ability to provide smooth trajectories and precise control over velocity and acceleration. The general form of a quintic polynomial is given by:

\begin{equation}
\label{eq:z_height}
f(t)=\sum_{k \leq 5} a_k t^k
\end{equation}

Let \( t \) denote the time variable, and \( a_{0}, a_{1}, ..., a_{5} \) be the coefficients that need to be determined. We denote the swing time by \( T \). In our periodic reward design, one leg being in the swing phase implies the other is in the stance phase. One swing phase and one stance phase together complete a full gait cycle. The trajectory of the robot's foot during the swing phase is defined by \( f(t) \), which is shaped through a series of kinematic constraints at critical moments in the robot's gait. These constraints are:

\begin{enumerate}
    \item Initial foot height at \( t = 0 \), given by \( f(0) = h_0 \), where \( h_0 \) is the initial height.
    \item Initial foot velocity at \( t = 0 \), determined by \( f'(0) = v_0 \), with \( v_0 \) being the initial velocity.
    \item Initial foot acceleration, represented by \( f''(0) = \text{acc}_0 \), where \( \text{acc}_0 \) is the initial acceleration.
    \item Reaching maximum foot height at the midpoint of the swing phase, \( f(T/2) = h_{\text{max}} \), where \( h_{\text{max}} \) is the target feet height.
    \item Final foot height at the end of the swing phase, \( f(T) = h_{\text{swing}} \), with \( h_{\text{swing}} \) as the final height.
    \item Final foot velocity at the end of the swing phase, \( f'(T) = v_{\text{swing}} \), where \( v_{\text{swing}} \) is the final velocity.
\end{enumerate}

To deduce the coefficients \( a_{0...5} \), a numerical optimization technique is employed. Once the coefficients are ascertained, they succinctly characterize the foot's vertical trajectory (i.e., the swing height). Quintic polynomial interpolation is instrumental in ensuring soft landings within humanoid robotic locomotion, offering granular control over the robot's swing height, foot acceleration, and velocity profiles. One optimization result and the corresponding trajectory plot are shown in Appendix Table \ref{tab:foot_trajectory_optimization} and Fig. \ref{fig:foot_traj}.

This method facilitates the manipulation of higher-order derivatives to attenuate impact forces at footfall. By adjusting the coefficients of the quintic polynomial, trajectories are crafted that not only elevate the foot to surmount obstacles but also maintain smooth movements and mitigate the impact upon contact. These fluent transitions enable a gentler touchdown, enhancing the robot's stability and advancing the creation of efficient, adaptable robots capable of safely traversing diverse terrains.

\subsection{Configuration of DWL training process}

In our DWL framework, as illustrated in Fig. \ref{fig:method}, we utilize a Gated Recurrent Unit (GRU)\cite{dey2017gate} for the encoding process and a two-layer Multilayer Perceptron for both the decoding and actor networks. Details of the training configuration can be found in the appendix section \ref{sec:train_config}.

The robot's base pose is denoted by \(P^b\), and the pose of the feet is denoted by \(P^f\). The pose, which includes both position and orientation, is represented as a six-dimensional vector \([x, y, z, \alpha, \beta, \gamma]\). Here, \(x, y, z\) specifies the position, and \(\alpha, \beta, \gamma\) represents the orientation in Euler angles. The policy network inputs include proprioceptive sensor data and a periodic clock signal, represented as \((\sin(t), \cos(t))\), in addition to command inputs defining the desired velocities \(\dot{P}_{x, y, \gamma}\). These observations are detailed in Table \ref{tab:observation}. The state includes privileged observations, which are typically unavailable to standard proprioceptive sensors on physical robots. This state also integrates the current step's reward, combining a reward model with the world model, which is expected to enhance the encoder's ability to capture the environmental context in the latent space.

Other important components of the state are the Periodic Stance Mask \(I(t)\), a binary indicator of expected foot contact patterns for a periodic gait, and the Cycle Time, essential for computing foot trajectories as outlined in \eqref{eq:z_height}. The Feet Movement, indicating both the position $P^f_{xyz}$ and velocity of the feet $\dot{P}^f_{xyz}$. Also, the height scan provides an approximate height map to further enhance the estimation of the state. Please note that the input to our policy includes only proprioceptive sensor data and does not incorporate any LiDAR or depth camera information. The height scans listed in Table~\ref{tab:observation} are privileged observations employed by the Critic during training.

\begin{table}[htp]
\centering
\caption{Summary of Observation Space. The table categorizes the components of the observation space into observation and state. The table also details their dimensions.}
\label{tab:observation}
\begin{tabular}{lccc}
\toprule
\textbf{Components} & \textbf{Dims} & \textbf{Observation} & \textbf{State} \\
\hline
Clock Input & 2 & \checkmark & \checkmark \\
Commands & 3 & \checkmark & \checkmark \\
Joint Position & 12 & \checkmark & \checkmark \\
Joint Velocity & 12 & \checkmark & \checkmark \\
Angular Velocity  & 3 & \checkmark & \checkmark \\
Orientation  & 3 & \checkmark & \checkmark \\
Last Actions & 12 & \checkmark & \checkmark \\
\hline
Base Linear Velocity & 3 & & \checkmark \\
Frictions & 1 & & \checkmark \\
Push Force\&Torques & 6 & & \checkmark \\
Cycle Time & 1 & & \checkmark \\
Periodic Stance Mask & 2 & & \checkmark \\
Feet movement & 12 & & \checkmark \\
Feet Contact & 2 & & \checkmark \\
Body Mass & 1 & & \checkmark \\
Current Reward & 1 & & \checkmark \\
Torques & 12 & & \checkmark \\
% Tracking Difference & 12 & & \checkmark \\
Height Scan & 96 & & \checkmark \\
\bottomrule
\end{tabular}
\end{table}

\begin{table}[htp]
\centering

\caption{Overview of Domain Randomization. Presented are the domain randomization terms and the associated parameter ranges. Additive randomization increments the parameter by a value within the specified range while scaling randomization adjusts it by a multiplicative factor from the same range.}

\label{tab:domain_randomization}
\begin{tabular}{lccc}
% \hline
\toprule
\textbf{Parameter} & \textbf{Unit} & \textbf{Range} & \textbf{Operator} \\
\hline
Joint Position & rad & [-0.3, 0.3] & additive \\
Joint Velocity & rad/s & [-1, 1] & additive \\
% Base Lin. Vel. & m/s & [0.0, 0.15] & additive \\
Angular Velocity & rad/s & [-0.1, 0.1] & additive \\
Orientation & rad & [-0.1, 0.1] & additive \\
System Delay & ms & [0, 10] & - \\
Friction & - & [0.2, 2.0] & - \\
% Action Delay & - & [0, 1] & - \\
Motor Offset & rad & [-0.05, 0.05] & additive \\
Motor Strength & \% & [90, 110] & scaling \\
% Motor Damping & Nms/rad & [0.3, 4.0] & scaling \\
Payload & kg & [-5, 20] & additive \\
PD Factors & \% & [80, 120] & scaling \\
% Kd Factor & \% & [0.9, 1.1] & scaling \\
% Gravity & m/s\textsuperscript{2} & [0.0, 0.67] & additive \\
% Restitution & - & [0.0, 0.4] & additive \\
\bottomrule
\end{tabular}
\end{table}

Each action \( a_t \in \mathbb{R}^{12} \) determines the actuators' target positions, followed by a Proportional-Derivative controller to translate into joint torques. Our control policy functions at 100Hz, surpassing the usual rates in RL-based locomotion strategies (50Hz), thus providing finer granularity and enhanced precision in the robot's movements. The internal PD controller operates at a higher frequency of 500Hz.

For our simulations, we use the Isaac Gym environment \cite{makoviychuk2021isaac}. However, its lack of support for the closed kinematic chain employed in our ankle control necessitates the addition of two virtual motors within the simulator. We then remap the joint targets to the actual motors for deployment. The policy optimization employs the DWL loss function (refer to Equation \ref{eq:dwl}) with Adam optimizer\cite{kingma2014adam}. This method capitalizes on the inherent advantages of the DWL framework to refine a robust locomotion policy, and Hyper-parameters can be found in Appendix TABLE \ref{tab:hyperparameters}. Remarkably, the resulting policy is ready for direct application to the physical robot without the need for further adjustments, exemplifying a seamless zero-shot transfer from simulation to real-world deployment.

\subsection{Domain Randomization}
\label{sec:DR}

To bridge the gap between simulation and reality, our methodology emphasizes extensive domain randomization of crucial dynamics parameters. This addresses the main sources of real-world variability: environment noise, dynamics noise, and sensory noise.

Randomization covers environmental elements such as floor friction, orientation, and robot-specific aspects like mass and Center of Mass positioning. Variations in motor parameters, including PD controller settings, are introduced to acclimate the policy to a range of motor behaviors.

Furthermore, we incorporate system latencies and inject random deviations in the robot's Center of Mass, equipping the policy to handle unforeseen disturbances in real environments. This thorough randomization strategy is essential for ensuring the policy's resilience and flexibility in actual deployment scenarios. Further specifics are presented in Table \ref{tab:domain_randomization}.

\begin{figure}[tp]
    \centering
    \includegraphics[trim={0.0cm 0.0cm 0.0cm 0.7cm},clip,width=\linewidth]{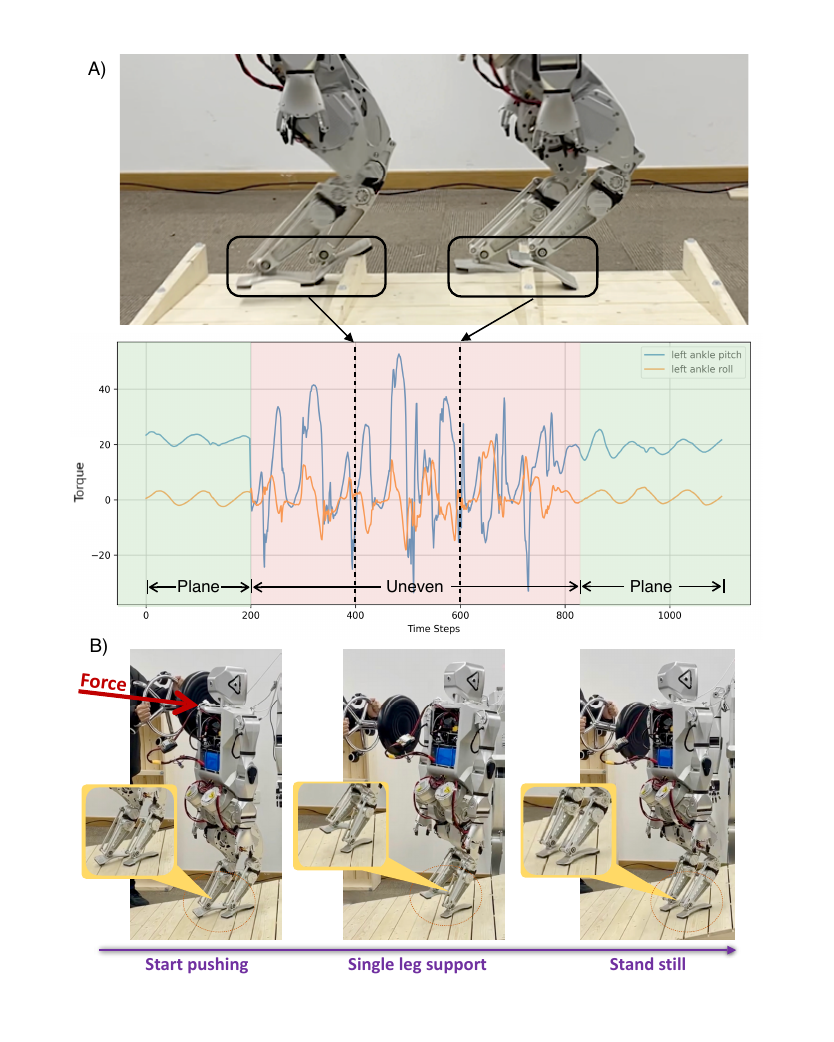}
    \vspace{-7mm}
    \caption{Dynamic adaptation of the ankle control mechanism. A) The top image demonstrates the humanoid robot's ankle control system actively maintaining balance on uneven terrain. The associated torque plot reveals the control system's adjustments during steady locomotion. B) The bottom image shows the system's resilience to external perturbations during static standing, where the 2-DoF ankle control plays a key role in maintaining stability.}
    \label{fig:ankle_control}
    \vspace{-0.15in}
\end{figure}
\section{Experiments}
\label{sec:exp}
In this section, we mainly focus on the performance of challenging settings in both indoor and outdoor environments. The benchmark comparisons discussed below were all conducted using the smaller humanoid robot, which stands 1.2 meters tall. Additionally, we deployed our algorithm on a larger humanoid robot, measuring 1.65 meters in height, as detailed in Fig.~\ref{fig:main_exp_L}.

\subsection{Benchmark Comparison}
For the empirical assessment of our approach, we conduct a series of experiments using both the XBot-S and XBot-L, applying the learned policy in a zero-shot transfer to real-world settings. This deployment encompasses a range of intricate and challenging terrains, testing the limits of the locomotive capabilities of our robot. To the best of our knowledge, this represents the world first humanoid robot to robustly navigate such complex environments using end-to-end reinforcement learning.

Our evaluation framework includes comparisons with two baseline methodologies, providing a comprehensive perspective on the effectiveness and advancements offered by our approach. In summary, we run three kinds of algorithms:

\begin{itemize}
    \item \textbf{DWL Baseline(ours):} This baseline involves the application of the DWL policy with active ankle control. For this configuration, we set the PD gains of the ankle joints to \(K_p=20\) and \(K_d=5\). The network architecture details of DWL can be found in Appendix Table~\ref{tab:network_architecture}. The total trainable parameters of the DWL actor is about $\textbf{320,192}$.
    \item \textbf{PPO with Ankle Control:} Here, we eliminate the denoising loss component while retaining the other aspects of our methodology. This setup aims to underscore the enhanced adaptability of our approach in contrast to traditional methods. The network architecture details of PPO can be found in Appendix Table~\ref{tab:ppo_network_architecture}. The total trainable parameters of the PPO actor is about $\textbf{333,312}$.
    \item \textbf{DWL without Ankle Control:} Given the complexity of modeling closed kinematic chain ankle mechanisms in bipedal and humanoid robots, many previous RL-based locomotion controls have utilized passive ankle strategies. We conduct comparisons with a DWL variant employing \textbf{passive ankle control}\footnote{We specify that ``passive" ankle control in our context means the ankle's movement isn't directly controlled by the policy but responds based on predefined physical damping properties, which is different from active RL agent decision-making.} (with \(K_p=0\) and \(K_d=10\)) to benchmark against this common approach.
    
\end{itemize}

Our experimental scenarios are diverse, including tasks such as snowy ground, up and down stairs, and disturbance rejection. During these tasks, the robot's arms are maintained in a stationary position to isolate the assessment of locomotive performance. This experimental setup provides a testbed to evaluate the versatility and robustness of our locomotion control strategy in real-world conditions. The subsequent benchmark comparisons are conducted exclusively using XBot-S to ensure consistency in our evaluation.
\begin{table}[htbp]
\centering
\caption{Real robot testing across various terrains. Bold values is our DWL with ankle control, $\text{DWL}_p$ is DWL with passive ankle, PPO control ankle as well\textcolor{blue}{.}}
\label{tab:robustness_test}
\begin{tabular}{lcccc}
% \hline
\toprule

\textbf{Algorithm} & \textbf{Slope} & \textbf{Stair-up} & \textbf{Stair-down} & \textbf{Irregular} \\
\hline
PPO & 80\% & 20\%  & 60\%  & 20\%  \\
$\text{DWL}_p$ & 80\% & 20\%  & 100\%  & 40\%  \\
\textbf{DWL} & 100\% & 100\% & 100\% & 100\% \\
\bottomrule
\end{tabular}
\end{table}

\subsection{Indoor Experimental Validation}
A comprehensive suite of real-world trials is executed to assess the robustness and adaptability of our algorithm in controlling the humanoid robot across a series of challenging terrains. Our indoor experiments employed four distinct terrain types with different difficulties, detailed as follows:

\begin{itemize}
    \item \textbf{Slope Transit (Fig. \ref{fig:main_exp}F):} A sloped platform with a gradient of 0.25 to test the robot's capacity to adeptly shift from planar to inclined locomotion, encompassing both ascent and descent.
    \item \textbf{Stair Descent (Fig. \ref{fig:main_exp}D):} Tasked with a downward traversal, the robot encountered stairs\textcolor{blue}{,} each spanning $20cm$ wide and $10cm$ high, commencing from the summit.
    \item \textbf{Stair Ascent (Fig. \ref{fig:main_exp}B):} Matching the descent staircase in dimension, the robot faces the challenge of ascending the stairs with limited sensor observations.
    \item \textbf{Irregular Terrain (Fig. \ref{fig:main_exp}E):} A custom-designed landscape with variable elevations up to $10cm$, simulating the unpredictability of challenging terrains.
\end{itemize}

The results of these experiments, quantified in Table \ref{tab:robustness_test}, reveal significant insights. 
% 从算法表格对比
When comparing PPO with ankle control to our DWL framework, our approach shows a significant improvement in walking performance, achieving \textbf{100\%} success rates across various terrains. This highlights the superior sim-to-real capabilities of our method and its robust adaptability to diverse terrains. For relatively simple tasks like navigating slopes and descending stairs, both PPO and the passive variant of DWL ($\text{DWL}_p$) demonstrate competent success rates. However, in challenging situations such as walking on irregular terrain or climbing stairs, DWL distinctly outperforms, showcasing the adaptability and robustness of our method.

% The importance of ankle control is equally evident when comparing whether DWL utilizes ankle control. From Table.\ref{tab:robustness_test}, it is clear that the inclusion of activate ankle control leads to an improvement in success rates especially in hard terrain. The challenging stair ascent particularly underscores the necessity of active ankle control, since this task requires for sustained single-leg stability throughout an extended swing phase,  highlighting the value of ankle control in preserving balance. This capability was pivotal for the humanoid robot's successful navigation across diverse terrains.

% 从任务出发对比
% For relatively straightforward terrains such as slopes and stair descents, both PPO and the passive variant of DWL ($\text{DWL}_p$) displayed competent success rates. In contrast, while walking on irregular terrain or climbing stairs, necessitated a higher degree of algorithmic adaptability, where DWL exhibited pronounced superiority.

% The challenging stair ascent particularly underscored the necessity of active ankle control. The task's requirement for sustained single-leg stability throughout an extended swing phase, due to stair climbing, foregrounds the value of ankle control in preserving balance. This capability was pivotal for the humanoid robot's successful navigation across diverse terrains, as demonstrated by the DWL algorithm's perfect success rates.

\subsection{Outdoor Experiments}

In addition to extensive indoor testing on various complex terrains, we also conducted prolonged outdoor walking tests across diverse and challenging landscapes. We assessed walking performance on different surfaces and conditions, including cemented ground, brick roads, soil, and snowy terrain. 
Our algorithm allowed the robot to exhibit stable walking across the aforementioned varied road conditions. 
Particularly noteworthy is the walking evaluation on snow-covered terrain Fig. \ref{fig:main_exp}A , which presents a highly challenging task. Snow, with its deformable nature, poses difficulties as the robot's feet can sink into it, a situation challenging to simulate. Furthermore, snow surfaces tend to be slippery, making robots prone to sliding. Our DWL algorithm, however, demonstrates remarkable stability during prolonged walks on snowy terrain, affirming the robustness and adaptability of our algorithm to diverse terrains.
\subsection{Robustness Testing}

Domain randomization is a common approach to achieving a robust controller. However, its effectiveness is often limited by the sim-to-real gap and the impracticality of accounting for every possible real-world scenario.

In our framework, the agent is designed to predict the true state, thereby empowering the controller to rapidly adapt to a variety of situations. For example, if there is a tendency to fall, the controller can quickly recognize this and act to maintain balance. To assess the robustness of our controller, we conducted the following experiments:

\subsubsection{Mass Displacement}
As shown in Fig. \ref{fig:main_exp}G, the robot carried a bag into which we progressively added heavy objects while it was walking. Even with an additional weight of up to 15kg, over a third of the robot's weight, it managed to sustain stable locomotion.

\subsubsection{Heavy load}
We executed two experiments, depicted in Fig. \ref{fig:main_exp}H. It successfully walked with an additional load of up to 20kg, albeit with a slightly lowered height due to the added weight. Also, we attached the robot’s hand effector to a loaded cart Fig. \ref{fig:main_exp}I, the robot was able to push a cart loaded with 60kg, demonstrating the controller's adaptability to handle significant loads.

\subsubsection{Push Recovery}

During our experimentation, we subjected the robot to external forces from multiple directions while it was executing continuous standing commands. These tests were carried out on both flat(Fig. \ref{fig:main_exp}C) and sloped terrain(Fig. \ref{fig:ankle_control}), with the robot successfully maintaining its stance in both conditions. 
% Here, the role of ankle control was pivotal, as it allowed the robot to preserve balance, even in single-leg support scenarios, mirroring human-like stability.

% \section{Result Analysis}
\begin{figure}[tp]
    \centering
    \includegraphics[width=1.0\linewidth]{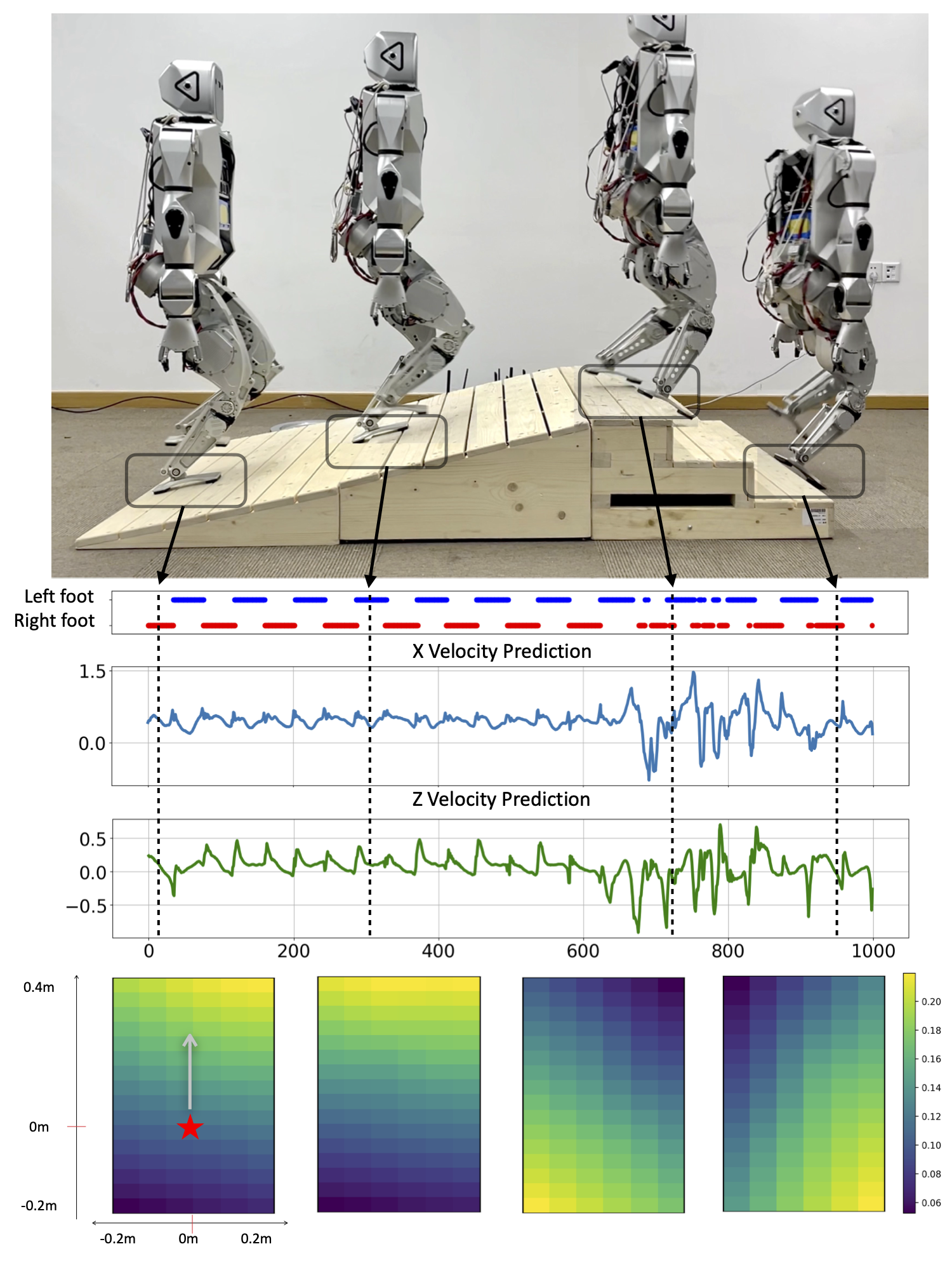}
    \caption{State estimation results of DWL-facilitated complex terrain traversing and adaptation. This sequence of images visualizes the model's prediction of foot contact, base velocity, and heightmap when the humanoid robot navigates through a slope and stairs. The results demonstrate DWL's effectiveness in state estimation and online adaption.}
    \label{fig:traj}
    \vspace{-0.15in}
\end{figure}

\section{Result Analysis}

In this section, we delve into the empirical results obtained from the deployment of the DWL framework on a humanoid robot, specifically focusing on its performance in terrain traversing. Our results are shown in Fig. \ref{fig:traj}.

\subsubsection{Terrain Height Scan Prediction and Gait Adaptation}

Our findings highlight the DWL's remarkable ability to predict terrain height, an essential factor for terrain adaptability. Utilizing only proprioceptive inputs, rather than relying on LiDAR or depth cameras, our system is capable of estimating a terrain's rough profile. At first glance, this might seem implausible, but our approach can discern the general trend of the terrain, as exemplified in Fig. \ref{fig:traj}. DWL's internal model adeptly encodes environmental features with distinct separability, thus facilitating accurate terrain recognition. The robot can identify whether it is walking up a slope or descending stairs. While precise shape prediction remains challenging, even a rough estimate proves immensely useful. This capability crucially affects the robot's gait, as observed when it transits from slope to stairs. Such gait modifications are imperative not just for navigating obstacles but also for ensuring balance and stability across diverse terrains.

% \subsubsection{Foot Contact Detection and Its Implications}
% The analysis of foot contact patterns reveals their dependency on the terrain type. In humanoid locomotion, especially during single-leg support phases, accurate foot contact detection is essential for stability. DWL aids in predicting these contact events, allowing for better planning of leg swing trajectories. The frequency and pattern of foot contacts change in response to state estimation, leading to gait adjustments on complex terrain.

\subsubsection{Foot Contact Detection and Its Implications}

As demonstrated in Fig. \ref{fig:traj}, foot contact patterns indicate a correlation with the type of terrain encountered. In humanoid locomotion, especially during single-leg support phases, accurate detection of foot contacts is essential for stability. The DWL framework aids in predicting these contact instances, thereby improving the planning of leg swing trajectories and facilitating effective obstacle avoidance. The frequency and pattern of foot contacts vary dynamically, guided by state estimation, leading to crucial gait adjustments and adaptations for complex terrain navigation.

% \subsubsection{Foot Avoidance Trajectory and Gait Analysis}
% We observe distinct gait patterns during stair ascent. The robot's foot reflexes adapt instantaneously to stumbling or slipping, stabilizing its posture and adjusting its gait accordingly. For example, when walking on the plane the robot’s body tilts closer to the ground, and also keeps a more straight upper body. Conversely, during stair ascent, the footstep height increases significantly, facilitating safe stair climbing.

\subsubsection{Velocity Estimation}
Velocity prediction, particularly linear velocity, is challenging to obtain directly from proprioceptive sensors but is critical for successful locomotion. DWL effectively estimates velocity states within a unified framework, addressing challenges like IMU angular yaw drift, as shown in Appendix Fig.~\ref{fig:IMU_drift}. This estimation enhances command following and prevents yaw deviations. The congruence between actual and estimated velocities observed in our experiments significantly aids in the robot's locomotion, ensuring smoother and more predictable movements. Since real states cannot be directly obtained in real-world settings, we validate our state prediction using a sim2sim transfer. We tested our policy in MuJoCo and compared the state estimation with the ground truth. The predictions of linear velocity and Euler yaw are displayed in Appendix Fig.\ref{fig:vel_pred} and Fig.~\ref{fig:euler_pred}.

The mean square error (MSE) of the forward velocity estimation in 60 seconds is $0.046$ in the sim-to-sim scenario, while the IMU drift was reduced by around 87\% in real-world experiments. These comparisons clearly demonstrate the accurate state estimation capabilities of the DWL algorithm.

\subsubsection{Benefits and Importance of Ankle Control}
\begin{figure}[tp]
    \centering
    \includegraphics[width=1.0\linewidth]{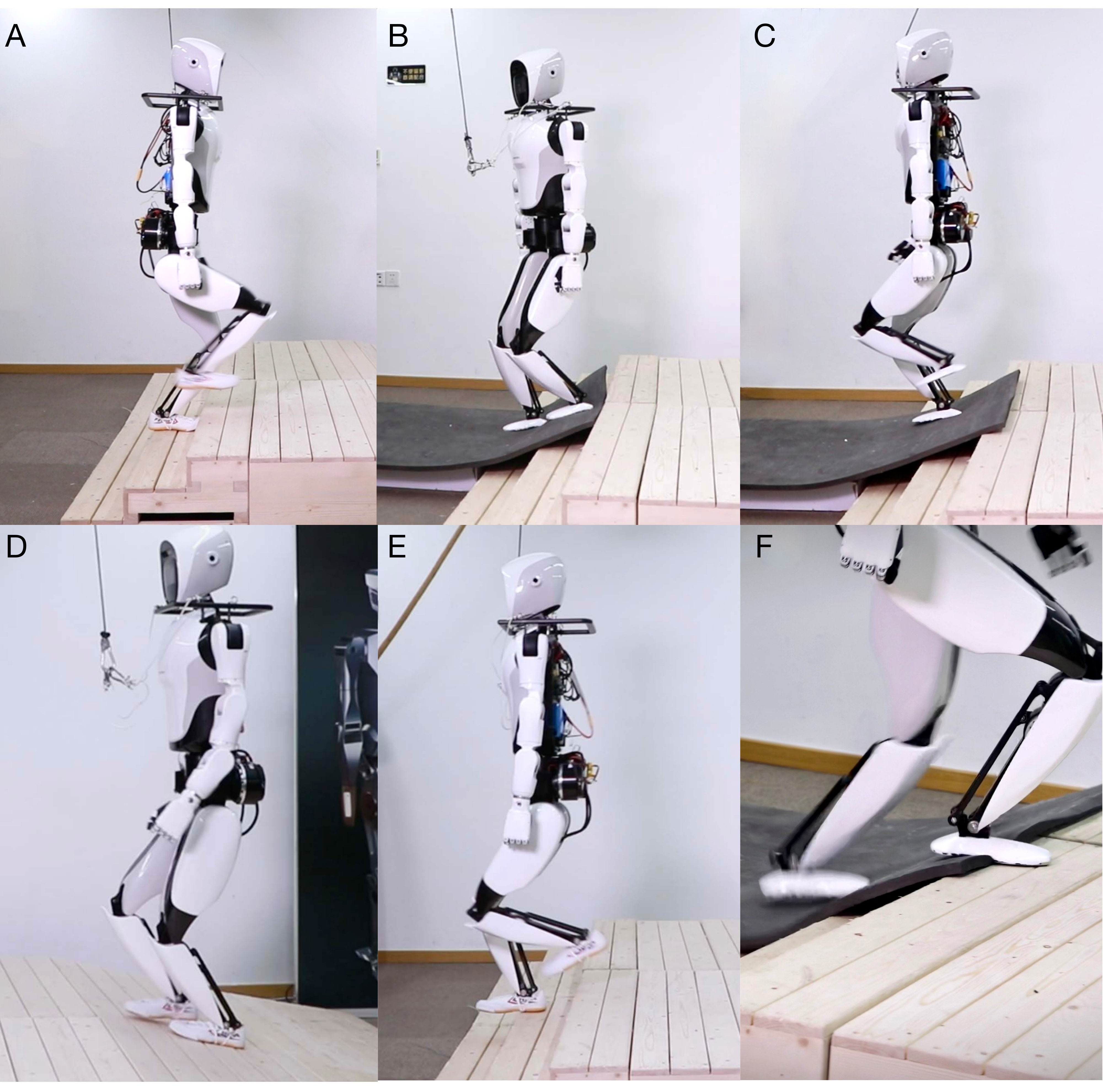}
    \caption{We deployed the DWL agent on a large humanoid robot (1.65 meters tall, 57 kilograms) to evaluate its performance in various scenarios, including highly challenging deformable conditions. The active control of the 2-DOF ankle joints proved particularly beneficial for maintaining balance while standing, recovering from external disturbances, and traversing through intricate and deformable terrain.}
    \label{fig:main_exp_L}
    \vspace{-0.15in}
\end{figure}

By allowing activated control of both the two freedoms of the ankle, the robot is empowered to traverse complex terrains and resist extra forces. The role of ankle control was pivotal, as it allows the robot to preserve balance, even in single-leg support scenarios, generating human-like stability.
Noticeably, ankle control becomes particularly pronounced when walking on irregular blocks and ascending stairs\textcolor{blue}{,} as in Fig. \ref{fig:ankle_control}, and also on deformable ground as shown in Fig.~\ref{fig:main_exp_L}. Without ankle control, the robot's feet deformed significantly upon contacting the uneven ground, failing to recover. This was coupled with inadequate ankle joint contact forces, raising the risk of imbalance and falls during foot placements. In contrast, our ankle control method navigates these terrains with ease by adaptive contact forces at the ankle joints\textcolor{blue}{,} as shown in the torque plot in Fig. \ref{fig:ankle_control}, enabling the robot's adaptability to varied terrains.

\section{Conclusion} 
\label{sec:conclusion}

In this work, we proposed Denoising World Model Learning (DWL) for complex humanoid robot locomotion skill learning. 
The framework first masked out privileged information and injected appropriate noise into the true state observed in the simulation. It then designs an auto-encoder architecture to denoise the observations and reconstruct the true state. We achieved success in extensive real-world experiments in various complex environments such as snowy land, stairs, deformable ground, and irregular surfaces. Demonstrating the world's first humanoid robot to master challenging terrains with end-to-end RL and zero-shot sim-to-real transfer. In-depth result analysis indicated the effectiveness of the state estimation capability and the importance of the active 2-DoF ankle control. In the future, visual information will be added to enable more efficient navigation in challenging terrains while maintaining robustness.

% In conclusion, this paper introduces the innovative Denoising World Model Learning framework, tailored for humanoid locomotion across challenging terrains. We have successfully deployed our policy on StarBot, a real humanoid robot, demonstrating its ability to traverse complex environments with remarkable agility and stability. The integration of the Parallel Double-Linkage Ankle Mechanism significantly enhances the robot's bipedal stability and maneuverability.

% The DWL framework efficiently overcomes the sim-to-real gap, facilitating a smooth transition from simulated environments to real-world applications without the need for extensive re-tuning. Its proficiency in end-to-end state representation learning and adaptation underscores its potential for zero-shot learning in robotic control systems.

% Our Denoising World Model Learning (DWL) approach and parallel double-linkage ankle control outperformed all the baselines in terms of performance and showed robustness in various challenging terrains in real-world settings. However, the current policy is blind. Therefore, in the future, we will aim to enhance the DWL framework by integrating visual inputs to achieve high performance and further deploying DWL to more challenging tasks, ultimately making robots human-level locomotion capabilities.
% Our ultimate goal is to advance humanoid robotics to a level where robots can seamlessly perform a diverse array of tasks in environments centered around human activities.

% \section*{Acknowledgments}

%% Use plainnat to work nicely with natbib. 

\newpage

\bibliographystyle{plainnat}
\bibliography{references}

\newpage

\section*{Appendix}
\subsection{Quintic Polynomial Foot Trajectory }
\begin{figure}[h]
    \centering
    \includegraphics[width=0.9\linewidth]{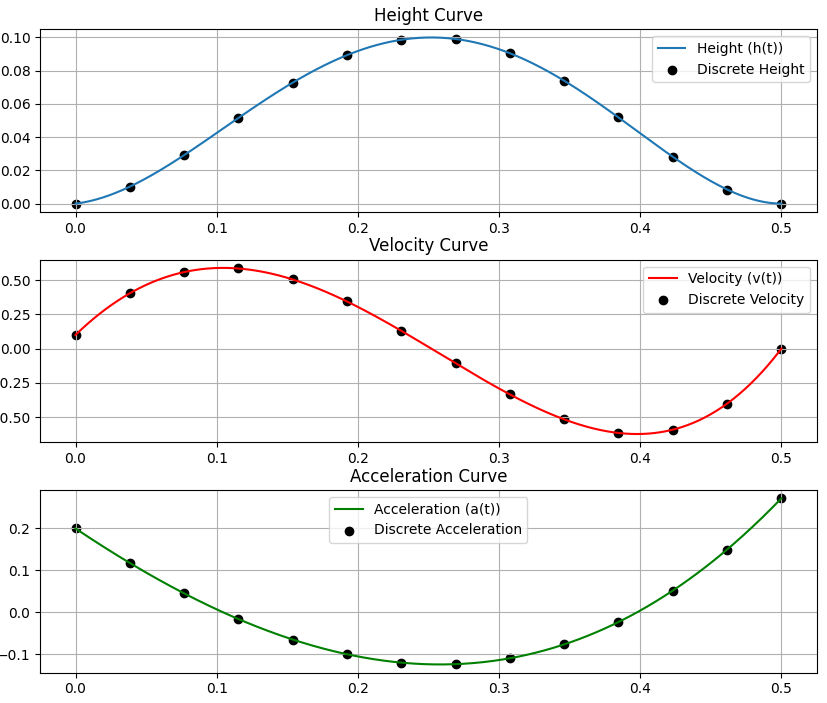}
    \caption{The foot trajectory, modeled through quintic polynomial interpolation, is detailed in TABLE \ref{tab:foot_trajectory_optimization}. It illustrates the critical velocity and acceleration constraints, ensuring a trajectory that facilitates seamless motion, harmonizing stability and gait efficiency in the robot's movement dynamics.}
    \label{fig:foot_traj}
\end{figure}

In Section.\ref{sec:Reward}, we introduced the foot trajectory interpolation method we designed. Through this approach, a smooth foot trajectory can be generated as a reference, and a tracking reward is utilized to encourage the robot's feet to follow the generated trajectory. Table.\ref{tab:foot_trajectory_optimization} presents the specific polynomial parameters and optimization conditions for generating foot trajectories. Figure.\ref{fig:foot_traj} illustrates the curves of height, velocity, and acceleration for a generated foot trajectory. It can be observed that smoothness is maintained across the positional, velocity, and acceleration aspects.
\begin{table}[h]
\centering
\caption{Quintic Polynomial Foot Trajectory Parameters $f(t)=\sum_{k \leq 5} a_k t^k$ and Optimization Conditions}
\label{tab:foot_trajectory_optimization}
\begin{tabular}{|c|c||c|c|}
\hline
\multicolumn{2}{|c||}{\textbf{Trajectory Parameters}} & \multicolumn{2}{c|}{\textbf{Optimization constraints}} \\
\hline
\textbf{Coefficient} & \textbf{Value} & \textbf{Objection} & \textbf{Value} \\
\hline
\(a_5\) & 9.6 & \(h_0\) & 0.0 \\
\(a_4\) & 12.0 & \(h_{T}\) & 0.0 \\
\(a_3\) & -18.8 & \(v_0\) & 0.1 \\
\(a_2\) & 5.0 & \(v_{T}\) & 0.0 \\
\(a_1\) & 0.1 & \(h_{max}\) & 0.1 \\
\(a_0\) & 0.0 & \(T\) & 0.5 \\
\hline
\end{tabular}
\end{table}

\subsection{Reward Function}
\label{sec:reward_funciton}
\begin{figure}[h]
    \centering
    \includegraphics[width=0.8\linewidth]{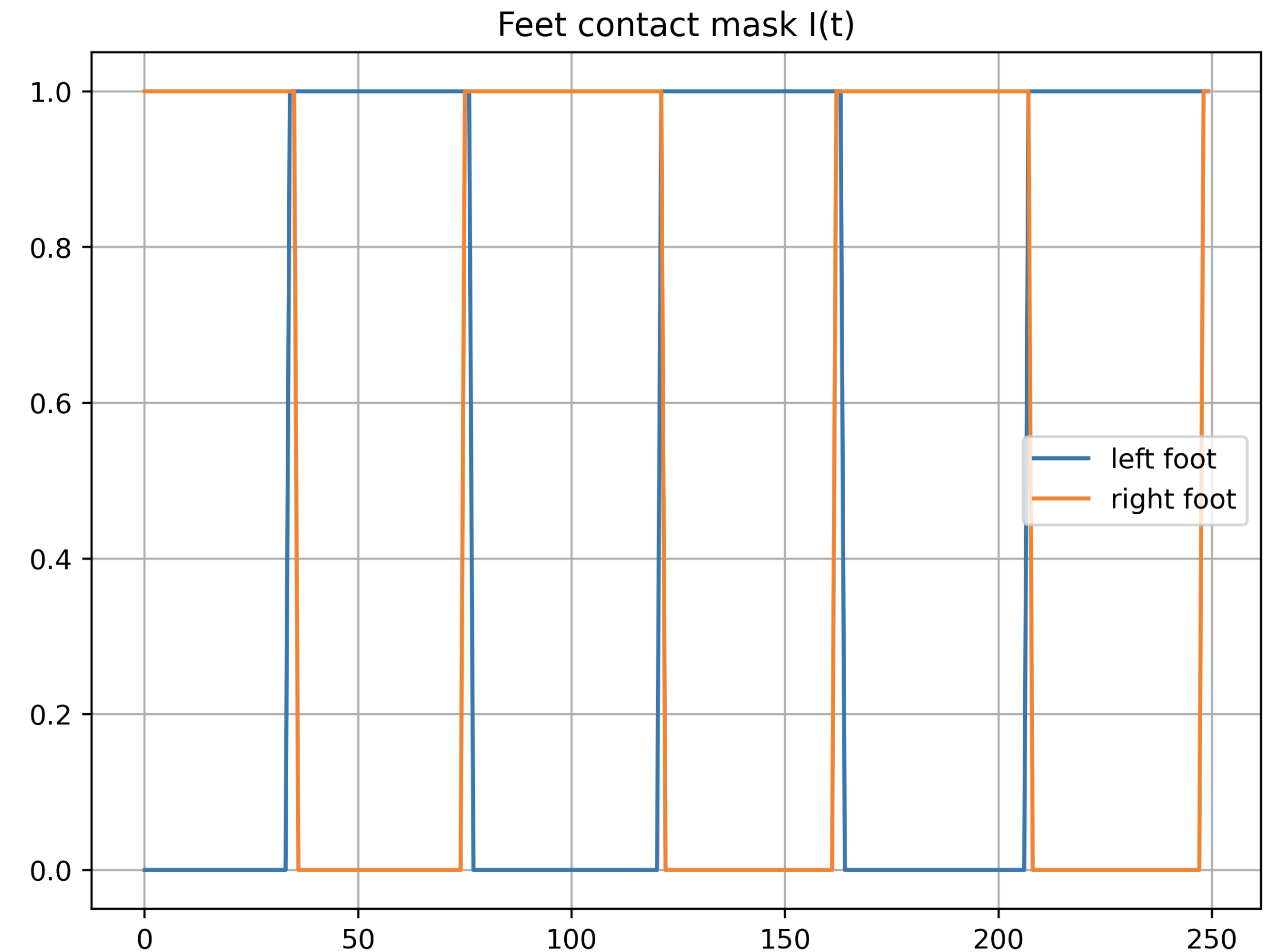}
    \caption{The stance mask for the left (L) and right (R) feet, where 0 indicates the swing phase and 1 indicates the stance phase is expected.}
    \label{fig:stance_mask}
\end{figure}

For periodic reward design, inspired by previous work \cite{siekmann2021sim, siekmann2021blind}, our objective is to construct a reward function that leverages the distinct roles of foot forces and foot velocities. Specifically, the function aims to promote higher foot velocities during the swing phase and reasonable foot forces during the stance phase of locomotion. We introduce a binary feet contact indicator, \( I(t) \), termed the periodic stance mask. As illustrated in Fig. \ref{fig:stance_mask}, this indicator is set to 1 during the planned contact phase and to 0 during the planned swing phase, alternating for each leg throughout a locomotion cycle. The periodic rewards are formulated as follows:

\begin{equation}
    r_{\text{Force}}^{\text{periodic}}(t) = I_L(t) \cdot F_L + I_R(t) \cdot F_R
\end{equation}

\begin{equation}
    r_{\text{velocity}}^{\text{periodic}}(t) = (1 - I_L(t)) \cdot \dot{P}^f_L + (1 - I_R(t)) \cdot \dot{P}^f_R
\end{equation}

In this context, the symbol \(F\) denotes force, while \(L\) and \(R\) refer to the left and right foot, respectively.
For clarity, we omit the scaling and clipping factors typically applied to the forces and velocities, which are necessary due to their differing magnitudes. For instance, forces, which can reach magnitudes of hundreds, are scaled down by a factor of 400 and subsequently clipped to the range [0, 1] to maintain consistency in the reward function's scale.

The reward function is summarized in Table \ref{tab:reward}. It is important to note that the commands $\text{CMD}_{z, \gamma, \beta}$ are intentionally set to zero. This is because we do not control them; rather, we aim to maintain their values at zero to ensure stable and smooth walking. Therefore, the total reward at any time step $t$ is computed as the weighted sum of individual reward components, expressed as $r_t = \sum_i r_i \cdot \mu_i $, where $\mu_i$ represents the weighting factor for each reward component $r_i$.

\begin{table}[hp]
\centering
\caption{In defining the reward function, we use a tracking error metric denoted by \(\phi(e, w)\). This metric is expressed as
\(
\phi(e, w) := \exp(-w \cdot \|e\|^2),
\)
where \(e\) represents the tracking error, and \(w\) is the associated weight. The target base height is set to \(0.7\,\text{m}\).
}
\label{tab:reward}

\begin{tabular}{@{}llr@{}}
\toprule
\textbf{Reward} & \textbf{Equation (\(r_i\))} & \textbf{reward scale(\(\mu_i\))} \\ \midrule

Lin. velocity tracking & $\phi(\dot{P}^b_{xyz} - \text{CMD}_{x y z}, 5)$ & 1.0 \\
Ang. velocity tracking &  $\phi(\dot{P}^b_{\alpha \beta \gamma} - \text{CMD}_{\alpha \beta \gamma}, 7)$ & 1.0 \\
% Linear velocity (\(z\)) & \(v_z^2\) & -2.0 \\
% Angular velocity (\(xy\)) & \(\omega_{xy}^2\) & -0.05 \\
Orientation tracking& \(\phi(P^b_{\alpha\beta}, 5)\) & 1.0 \\
% Joint accelerations & \(\dot{\theta}^2\) & \(-2.5 \times 10^{-7}\) \\
Base height tracking & $\phi(P^b_{z} - 0.7, 10)$ & 0.5 \\
\hline
Periodic Force & $r_{Force}^{periodic}(t)$ &  1.0 \\
Periodic Velocity & $r_{velocity}^{periodic}(t)$ &  1.0 \\
\hline
Foot height tracking  & $\phi(P^f_{z} - f_t, 5)$ & 1.0 \\
Foot vel tracking  & $\phi(\dot{P}^f_{z} - \dot{f}_t, 3)$ & 0.5 \\
% Action rate & \((a_t - a_{t-1})^2\) & -0.01 \\
\hline
Default Joint & $ \phi(\theta_t - \theta_0, 2) $ & 0.2 \\ 
Energy Cost & \(|\tau||\dot{\theta}|\) & -0.0001 \\
Action Smoothness & $ \| a_t - 2a_{t-1} + a_{t-2}\|_2$ & -0.01 \\
Feet movements & $ \| \dot{P}^f_{z} \|_2 + \| \ddot{P}^f_{z} \|_2$ & -0.01 \\
Large contact & $\text{CLIP}(F_{L,R} - 400, 0, 100)$ & -0.01 \\
% Power distribution & \(\text{var}(\tau \cdot \dot{\theta})^2\) & \(-10^{-5}\) \\
\bottomrule
\end{tabular}
\end{table}

\subsection{Training Details}
\label{sec:train_config}
Section\ref{sec:DWL} provides a detailed explanation of our designed Denoising World Model Learning(DWL) method, and here in Table.\ref{tab:hyperparameters}, we outline the hyperparameters for DWL. Table.\ref{tab:network_architecture} presents the specific network architecture we utilized. It includes the encoder and decoder for DWL, as well as the actor and critic.

\begin{table}[htbp]
\centering
\caption{DWL Network Architecture Details}
\label{tab:network_architecture}
\begin{tabular}{|c|c|}
\hline
\textbf{Component} & \textbf{Configuration} \\
% \hline
% Actor RNN & GRU(47, 256) \\
% \hline
% Critic RNN & GRU(301, 512) \\
\hline
% Latent Dimensions & 24 \\
% \hline
\multicolumn{2}{|c|}{\textbf{Encoder}} \\
\hline
RNN\_memory(0) & GRU(47 $\rightarrow$ 256) \\
emb\_model (1) & Linear(256 $\rightarrow$ 256) \\
emb\_model (2) & ELU(alpha=1.0) \\
emb\_model (3) & Linear(256 $\rightarrow$ 24) \\
\hline
\multicolumn{2}{|c|}{\textbf{Decoder}} \\
\hline
denoise\_net (0) & Linear(24 $\rightarrow$ 64) \\
denoise\_net (1) & ELU(alpha=1.0) \\
denoise\_net (2) & Linear(64 $\rightarrow$ 184) \\
\hline
\multicolumn{2}{|c|}{\textbf{Actor}} \\
\hline
policy\_net (0) & Linear(24 $\rightarrow$ 48) \\
policy\_net (1) & ELU(alpha=1.0) \\
policy\_net (2) & Linear(48 $\rightarrow$ 12) \\
\hline
\multicolumn{2}{|c|}{\textbf{Critic}} \\
\hline
Critic\_Net (0) & Linear(184 $\rightarrow$ 512) \\
Critic\_Net (1) & ELU(alpha=1.0) \\
Critic\_Net (2) & Linear(512 $\rightarrow$ 512) \\
Critic\_Net (3) & ELU(alpha=1.0) \\
Critic\_Net (4) & Linear(512 $\rightarrow$ 256) \\
Critic\_Net (5) & ELU(alpha=1.0) \\
Critic\_Net (6) & Linear(256 $\rightarrow$ 1) \\
\hline
\end{tabular}
\end{table}

\begin{table}[htbp]
\centering
\caption{PPO Network Architecture Details}
\label{tab:ppo_network_architecture}
\begin{tabular}{|c|c|}
\hline
\textbf{Component} & \textbf{Configuration} \\
\hline
\multicolumn{2}{|c|}{\textbf{Actor}} \\
\hline
RNN\_memory(0) & GRU(47 $\rightarrow$ 256) \\
% emb\_model (1) & Linear(256 $\rightarrow$ 256) \\
% emb\_model (3) & Linear(256 $\rightarrow$ 256) \\
% \hline
% \multicolumn{2}{|c|}{\textbf{Decoder}} \\
% \hline
% denoise\_net (0) & Linear(24 $\rightarrow$ 64) \\
% denoise\_net (1) & ELU(alpha=1.0) \\
% denoise\_net (2) & Linear(64 $\rightarrow$ 184) \\
% \hline
% \multicolumn{2}{|c|}{\textbf{Actor Policy}} \\
% \hline
% emb\_model (2) & ELU(alpha=1.0) \\
policy\_net (1) & Linear(256 $\rightarrow$ 256) \\
policy\_net (2) & ELU(alpha=1.0) \\
policy\_net (3) & Linear(256 $\rightarrow$ 128) \\
policy\_net (4) & ELU(alpha=1.0) \\
policy\_net (5) & Linear(128 $\rightarrow$ 12) \\
\hline
\multicolumn{2}{|c|}{\textbf{Critic}} \\
\hline
Critic\_Net (0) & Linear(184 $\rightarrow$ 512) \\
Critic\_Net (1) & ELU(alpha=1.0) \\
Critic\_Net (2) & Linear(512 $\rightarrow$ 512) \\
Critic\_Net (3) & ELU(alpha=1.0) \\
Critic\_Net (4) & Linear(512 $\rightarrow$ 256) \\
Critic\_Net (5) & ELU(alpha=1.0) \\
Critic\_Net (6) & Linear(256 $\rightarrow$ 1) \\
\hline
\end{tabular}
\end{table}

\begin{table}[htbp]
\centering
\caption{Hyperparameters of DWL.}
\label{tab:hyperparameters}
\begin{tabular}{lc}
\toprule
\textbf{Parameter} & \textbf{Value} \\
\hline
% Number of GPUs & 1 RTX A6000 \\
Number of Environments & 12288 \\
Number Training Epochs & 2 \\
% Steps per Environment & 49152\\
Batch size & $12288 \times 24$ \\
Episode Length & 2400 steps \\
Discount Factor & 0.995 \\
GAE discount factor & 0.95 \\
Entropy Regularization Coefficient & 0.005 \\
$c1$ & 0.8 \\
$c2$ & 1.2 \\
Learning rate & 1e-5 \\
regularization coefficient $\lambda_r$ & 0.002 \\ 
policy coefficient $\lambda_{\pi}$ & 5 \\ 
value coefficient $\lambda_v$ & 5 \\ 
% \hline
\bottomrule

\end{tabular}
\end{table}

\subsection{Additional Experimental Results}

\begin{figure}[htp]
    \centering
    \includegraphics[width=0.98\linewidth]{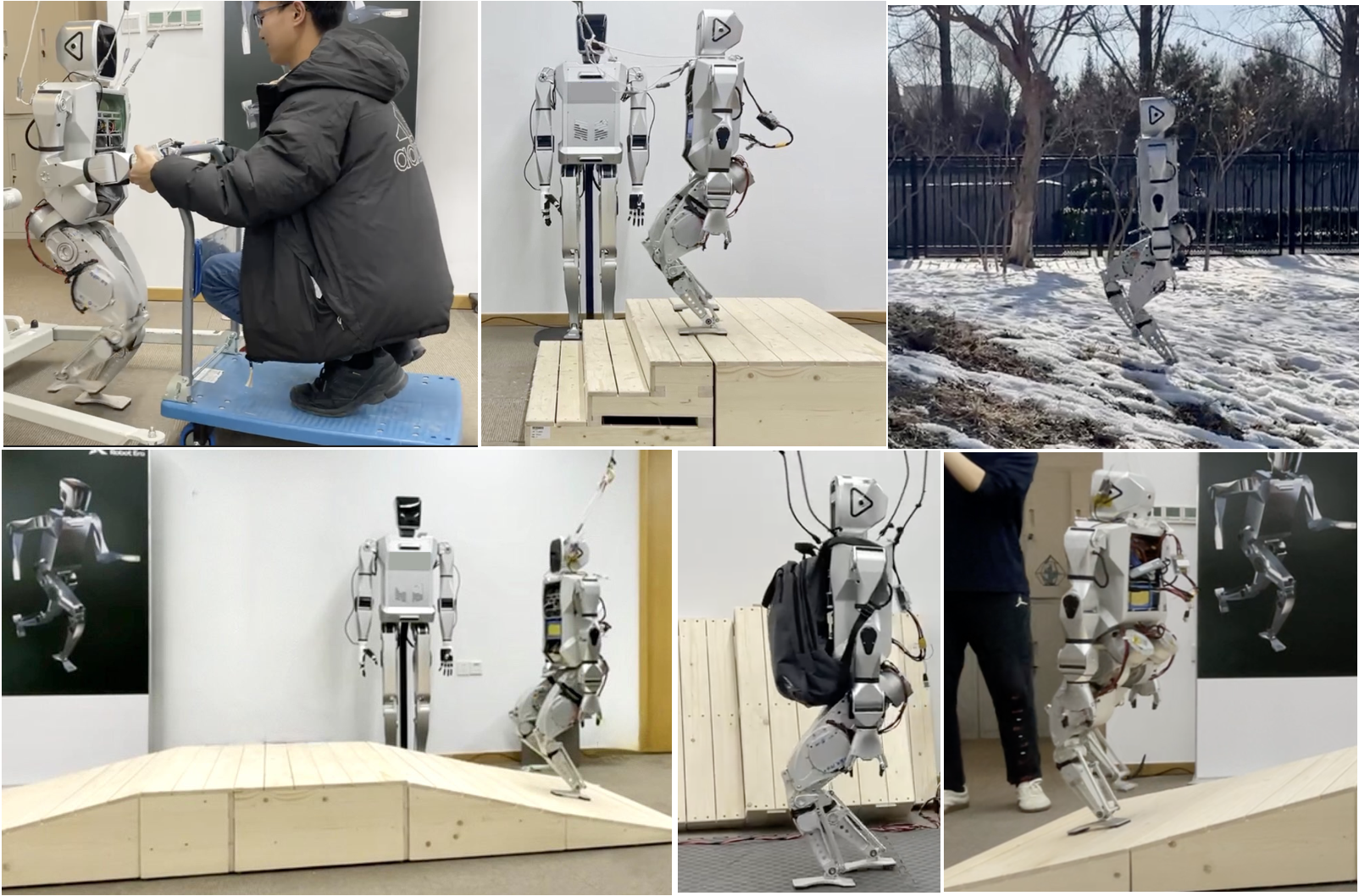}
    \caption{Additional Experiment Setups and Results}
    \label{fig:more_result}
\end{figure}

We present further results and demonstrations of our experiments, as illustrated in Fig. \ref{fig:more_result}.

% \begin{figure}[tp]
%     \centering
%     \includegraphics[width=1.0\linewidth]{images/main_exp_L.pdf}
%     \caption{\textcolor{blue}{We deployed DWL for testing on a large humanoid robot (1.65 meters tall, weighing 57 kilograms). The 2-DOF ankle is especially useful while standing and recovering from pushing and also on deformable ground.}}
%     \label{fig:main_exp_L}
%     \vspace{-0.15in}
% \end{figure}

\begin{figure}[tp]
    \centering
    \includegraphics[width=1.0\linewidth]{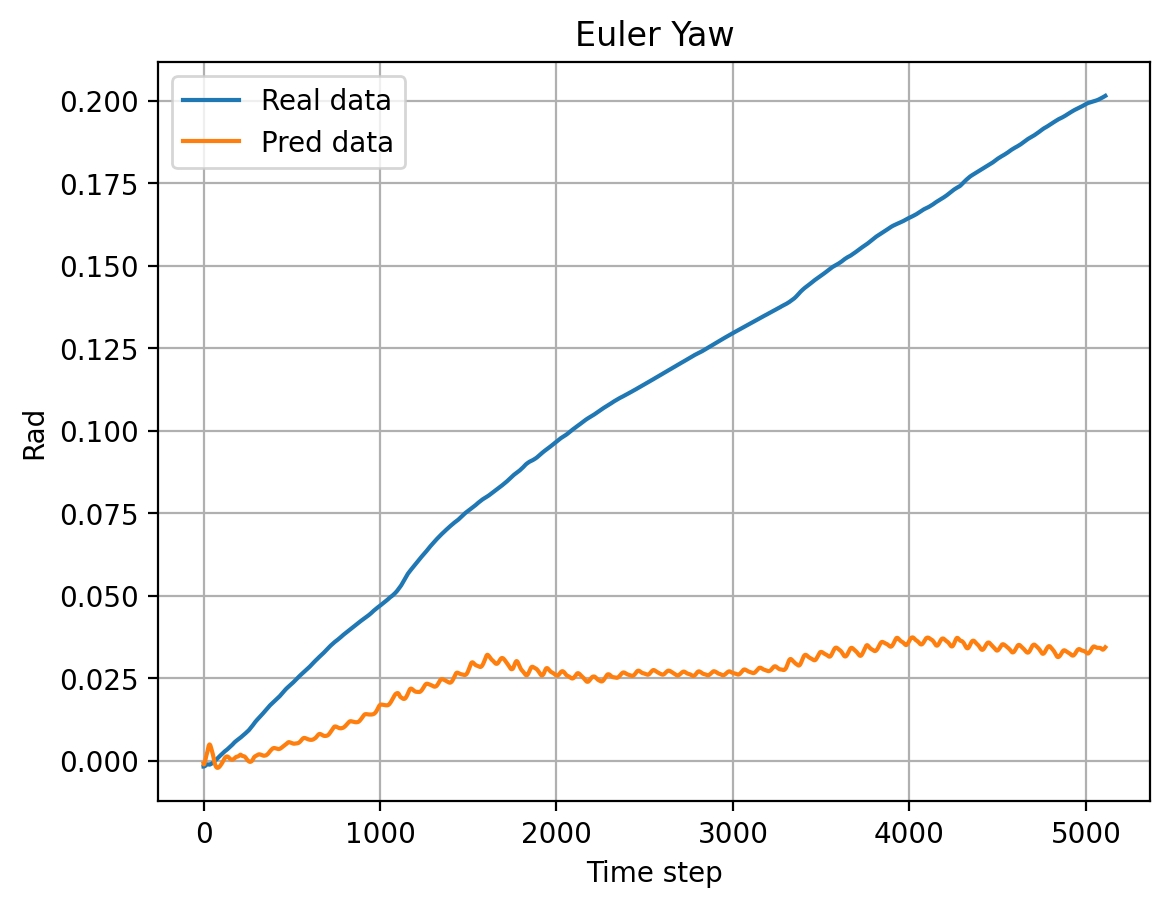}
    \caption{Even when the robot is stationary, the actual IMU readings exhibit the phenomenon of IMU drift. Conversely, the DWL algorithm is capable of predicting real IMU data and mitigating the IMU drift by approximately 87\%.}
    \label{fig:IMU_drift}
    \vspace{-0.15in}
\end{figure}

\begin{figure}[tp]
    \centering
    \includegraphics[width=1.0\linewidth]{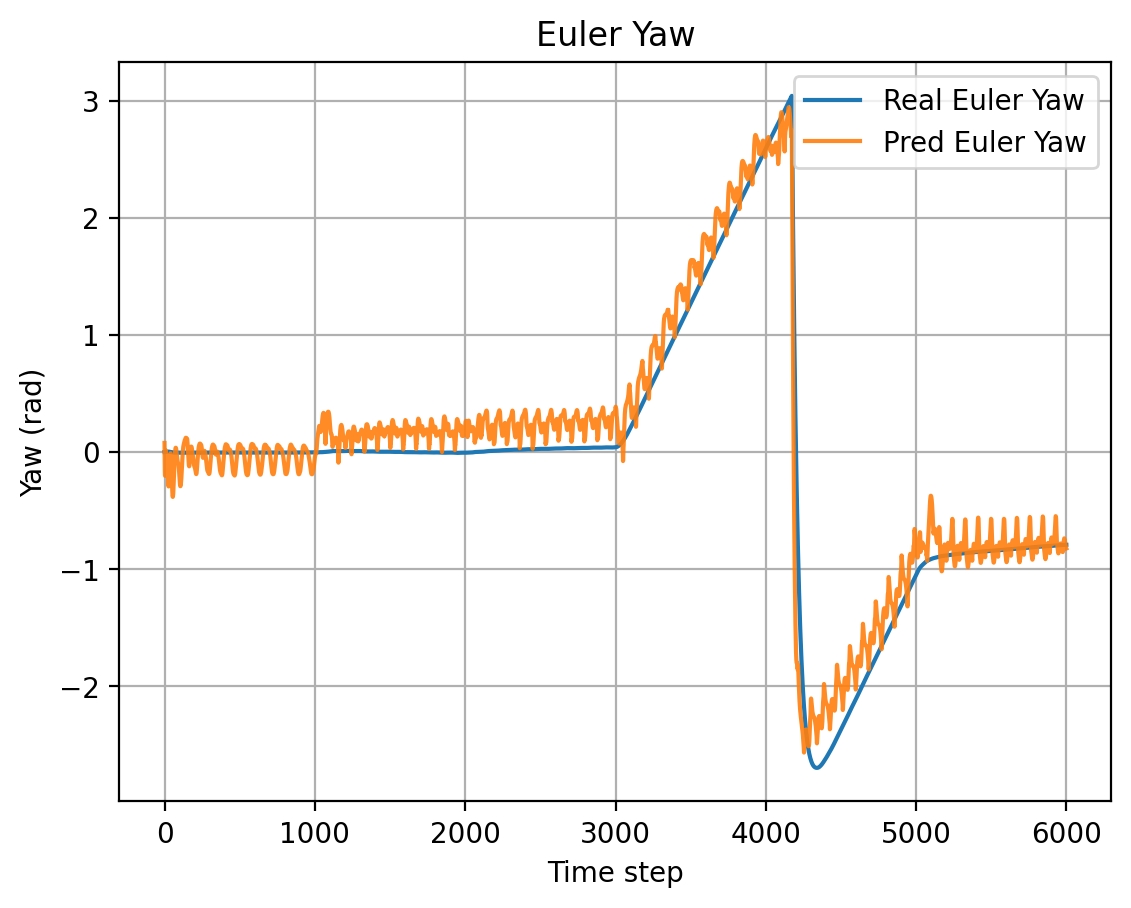}
    \caption{To emulate human command inputs, we varied the command input every 1000 steps (10 seconds). In the MuJoCo simulation environment, the Euler yaw angle predicted by the DWL algorithm is close to the ground truth with MSE $0.074$.}
    \label{fig:euler_pred}
    \vspace{-0.15in}
\end{figure}

\begin{figure}[tp]
    \centering
    \includegraphics[width=1.0\linewidth]{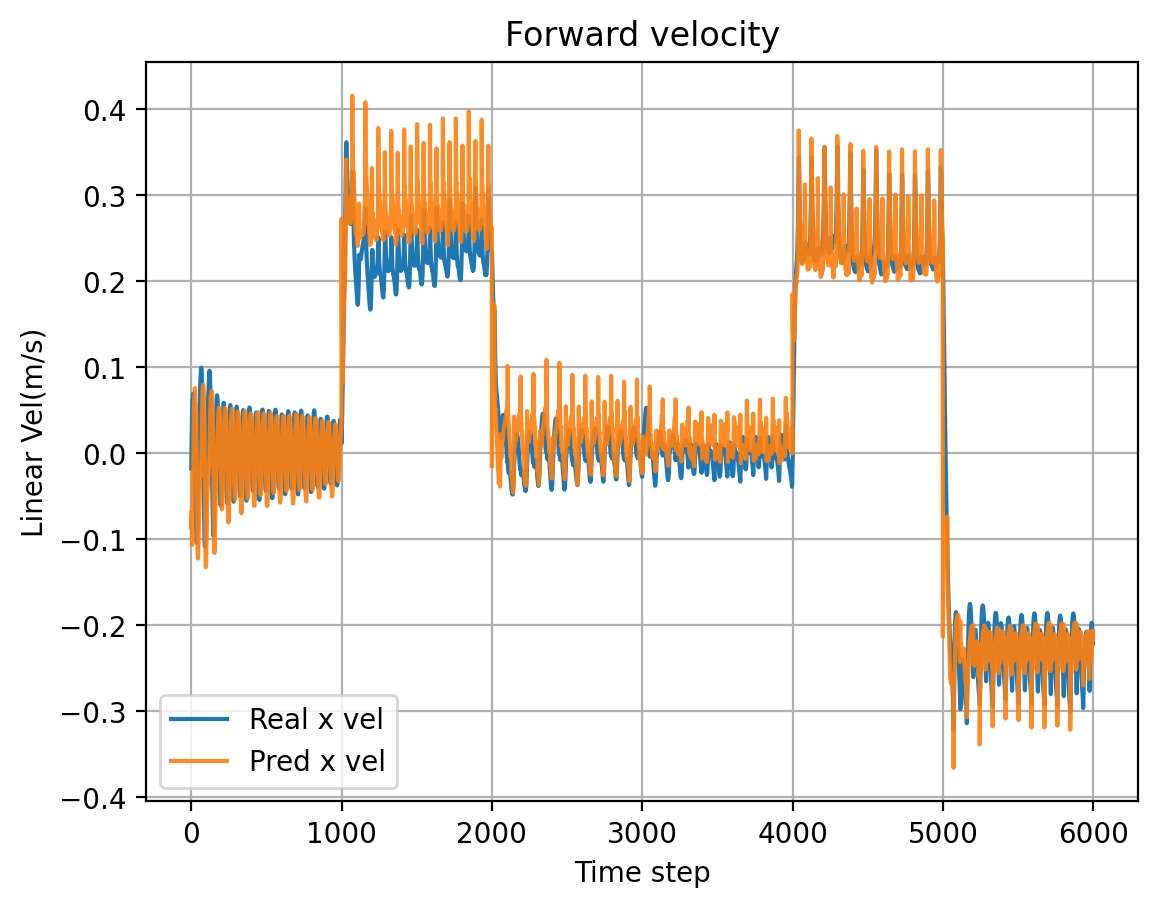}
    \caption{Comparison of Estimated and Actual Velocity Values in MuJoCo: The forward velocity predicted by the DWL algorithm closely approximates the ground truth with MSE $0.046$.}
    \label{fig:vel_pred}
    \vspace{-0.15in}
\end{figure}

\end{document}